\documentclass{article} 
\usepackage{iclr2027_conference,times}

\usepackage{amsmath,amsfonts,bm}

\def\eqref#1{equation~\ref{#1}}

\def\1{\bm{1}}

\DeclareMathAlphabet{\mathsfit}{\encodingdefault}{\sfdefault}{m}{sl}
\SetMathAlphabet{\mathsfit}{bold}{\encodingdefault}{\sfdefault}{bx}{n}

\usepackage{colortbl}
\usepackage{color}
\usepackage[dvipsnames]{xcolor} 
\usepackage[colorlinks=True,
            linkcolor=BrickRed,      
            anchorcolor=BrickRed, 
            citecolor=OliveGreen,   
            ]{hyperref}
\usepackage{url}
\usepackage{booktabs}
\usepackage{multirow}
\usepackage{graphicx}
\usepackage{makecell}
\usepackage{wrapfig}
\usepackage{enumitem}
\usepackage{xspace}
\usepackage{pifont}
\usepackage[table]{xcolor}
\usepackage{tabularx}
\usepackage{array}
\definecolor{oursbg}{HTML}{E5E7EB}

\newcommand{\themodel}{\textbf{\texttt{AIM}}\xspace}

\title{\themodel: Adaptive Interaction Modeling Networks for Real-to-Sim Soft-Body Simulation}

\author{
Tiancheng Yang$^{1,2,*}$,
Dingshuo Chen$^{1,3,*}$,
Tianle Chen$^{1,3}$,
Zhaocheng Liu$^{4}$,
Qiang Liu$^{1,3,\dagger}$
\\[0.3em]
$^1$NLPR, MAIS, Institute of Automation, Chinese Academy of Sciences\\
$^2$School of Advanced Interdisciplinary Sciences, University of Chinese Academy of Sciences\\
$^3$School of Artificial Intelligence, University of Chinese Academy of Sciences\\
$^4$Independent Researcher
}

\iclrfinalcopy 
\begin{document}

\maketitle

\fancyhead{}
\renewcommand{\headrulewidth}{0pt}

\begingroup
\renewcommand{\thefootnote}{\fnsymbol{footnote}}

\footnotetext[1]{The first two authors contributed equally to this work.}
\footnotetext[2]{Correspondence to Qiang Liu
(\texttt{qiang.liu@nlpr.ia.ac.cn}).}

\endgroup

\begin{abstract}
Deformable-object manipulation is essential for robotic tasks such as folding laundry and handling food, where robots must control shape changes as well as object motion. Predictive soft-body simulation supports these tasks by anticipating deformation under external interactions. However, spatial neighborhoods can misrepresent deformation dependencies, introducing local errors that accumulate over successive predictions. Models fitted to individual scenes must also accommodate changes in object geometry and manipulation conditions. In this work, we propose \themodel, an Adaptive Interaction Modeling framework that treats real-to-sim soft-body simulation as a local-global interaction modeling problem. \themodel uses motion history and geometry to adapt particle relations over current spatial neighbors and retained connections, while geometry-conditioned global communication coordinates object-wide responses. A unified kinematic control-point interface represents different manipulation configurations, and multi-step autoregressive supervision trains the model on its own predicted trajectories. Experiments on PhysTwin and PGND demonstrate improved motion accuracy and visual fidelity, with a 20.0\% reduction in future-prediction tracking error relative to PhysTwin and a 22.8\% reduction in mean long-horizon particle error across six object categories relative to PGND. The framework further supports transfer across actions, object instances, and scenes, including zero-shot transfer from robot interactions to human manipulation without target-domain dynamics fitting.
\end{abstract}

\section{Introduction}
\begin{figure}[t] 
\centering
\includegraphics[width=1.0\linewidth]{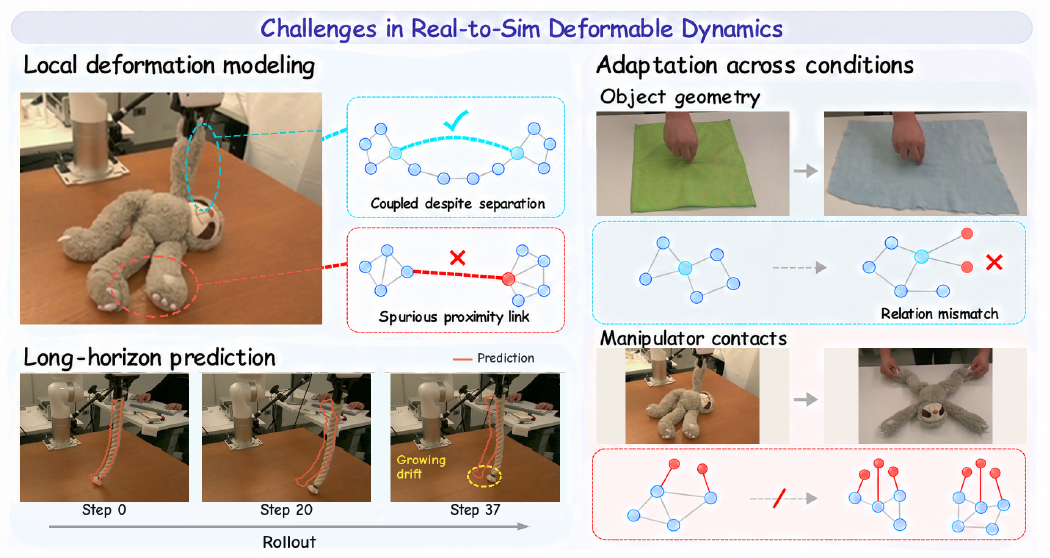}
\vspace{-14pt}
\caption{Challenges in real-to-sim deformable dynamics.}
\label{fig:main_fig}
\vspace{-12pt}
\end{figure}

Everyday tasks such as folding laundry, preparing food, and assisting with dressing require robots to control how objects deform under contact, making deformable-object manipulation an important capability for embodied intelligence~\citep{sanchez2018robotic,lin2021softgym}. Unlike rigid-object manipulation, these tasks involve spatially distributed responses to local actions. Predictive soft-body simulation enables robots to anticipate these responses and evaluate manipulation strategies before physical execution~\citep{lin2022learning,lin2021softgym,jiang2025phystwin}. While visual reconstruction captures observed deformation in increasing detail~\citep{pumarola2021d,park2021nerfies,luiten2024dynamic,yang2024deformable}, simulation must predict its evolution under external interactions. Particle- and mesh-based simulators provide a relational foundation through graph message passing~\citep{battaglia2016interaction,sanchez2020learning,pfaff2020learning}, and recent methods such as PhysTwin, PGND, and SoMA connect dynamics modeling with real observations and manipulation settings~\citep{jiang2025phystwin,zhang2025particle,huang2026soma}. Despite this progress, predicting object deformation remains challenging over long rollouts and under new interaction conditions, as illustrated in Fig.~\ref{fig:main_fig}.

Local deformation depends on structural coupling that spatial proximity alone cannot describe: folding brings structurally distant regions together, while stretching separates regions whose responses remain coupled~\citep{lin2022learning}. Misrepresenting these dependencies or their effects across the object introduces local motion errors~\citep{fortunato2022multiscale}, which can accumulate into long-horizon drift as autoregressive simulation repeatedly advances from its own predictions~\citep{um2020solver}. Addressing these modeling and prediction errors within a single scene is only part of the challenge, as practical manipulation also requires the simulator to generalize across objects and interactions. Changes in object geometry and manipulator contacts alter how actions induce deformation, challenging the generalization of learned interaction patterns beyond the training setting~\citep{zhang2024adaptigraph,shi2024robocraft}. Generalization therefore involves both accommodating the new configuration and preserving deformation accuracy throughout the resulting rollout.

In this work, we propose \themodel, an \underline{A}daptive \underline{I}nteraction \underline{M}odeling framework for real-to-sim soft-body simulation. \themodel combines history-conditioned adaptive particle relations with geometry-conditioned global context to model local deformation and coordinate object-wide responses. A unified kinematic control-point interface represents external interactions without action-category or manipulator-specific labels. The framework is trained through multi-step autoregressive prediction to learn dynamics from recursively generated states.

We evaluate \themodel on PhysTwin and PGND through resimulation, future prediction, and cross-condition transfer. Our method improves geometric and visual accuracy, reducing mean long-horizon MDE by 22.8\% relative to PGND across six object categories and future-prediction tracking error by 20.0\% relative to PhysTwin. We further demonstrate generalization across objects, actions, and scenes, including zero-shot transfer from robot interactions to human manipulation.

The main contributions of this work are three-fold:
\par\noindent\ding{182} We formulate real-to-sim soft-body simulation as an adaptive local-global interaction modeling problem, emphasizing how deformation-dependent particle relations and shared object-level context jointly support coherent dynamics under changing manipulation conditions.

\par\noindent\ding{183} We introduce \themodel, a particle-based framework that combines history-conditioned adaptive relations with geometry-conditioned global communication to coordinate local and object-wide responses under a unified kinematic control interface.

\par\noindent\ding{184} Experiments on PhysTwin and PGND demonstrate improved geometric and visual accuracy, with a 22.8\% reduction in mean long-horizon MDE over PGND across six object categories and a 20.0\% reduction in future-prediction tracking error over PhysTwin, alongside transfer across objects, actions, and manipulation domains.

\section{Related Work}
\label{sec:related_work}
\textbf{Real-to-Sim Deformable Dynamics.}
Recent real-to-sim methods connect visual observations with interactive simulation. PhysGaussian, Spring-Gaus, and PhysTwin couple visual representations with explicit physical models~\citep{xie2024physgaussian,zhong2024reconstruction,jiang2025phystwin}. Alternatively, GS-Dynamics, PGND, and SoMA learn action-conditioned state transitions from observed interactions~\citep{zhang2024dynamic,zhang2025particle,huang2026soma}. These approaches differ in their physical representations and transfer assumptions, ranging from new interactions with an identified object to unseen object instances. Our work follows the learned-dynamics direction, studying how interaction representations support both within-domain prediction and transfer across objects and manipulation conditions.

\textbf{Graph-Based Interaction Modeling.}
Interaction Networks, GNS, and MeshGraphNets model physical evolution through local relational computations~\citep{battaglia2016interaction,sanchez2020learning,pfaff2020learning}. Latent relation inference~\citep{kipf2018neural,graber2020dynamic} and multiscale propagation~\citep{fortunato2022multiscale,grigorev2023hood} address complementary questions of which elements should communicate and how information travels across an object. Building on the node--edge--global formulation~\citep{battaglia2018relational}, our framework couples history-conditioned relation learning with shared object context and geometry-conditioned particle updates. Appendix~\ref{app:related_work} discusses the broader physical, structural, and temporal modeling literature.

\section{Preliminaries}
\label{sec:preliminaries}

\textbf{Problem formulation.}
We represent a deformable object by $N$ particles $\mathcal{X}^{t}=\{\mathbf{x}_{i}^{t}\}_{i=1}^{N}$ and external manipulation by $M$ kinematic control points $\mathcal{C}^{t}=\{\mathbf{c}_{j}^{t}\}_{j=1}^{M}$, with positions in $\mathbb{R}^{3}$. The action $\mathcal{A}^{t}=\{(\mathbf{u}_{j}^{t},\boldsymbol{\alpha}_{j}^{t})\}_{j=1}^{M}$ contains prescribed displacements $\mathbf{u}_{j}^{t}=\mathbf{c}_{j}^{t+1}-\mathbf{c}_{j}^{t}$ and available geometric attributes. Given an $H$-frame history, the simulator predicts
\begin{equation}
    (\hat{\mathcal{X}}^{t+1},\mathcal{S}^{t+1})=F_{\theta}\big(\mathcal{X}^{t-H+1:t},\mathcal{C}^{t-H+1:t},\mathcal{A}^{t},\mathcal{S}^{t}\big),
    \label{eq:dynamics_formulation}
\end{equation}
where $\mathcal{S}^{t}$ stores retained particle pairs and their scores and is initially empty. Rollouts recursively reuse predicted object states under externally prescribed controls.

\textbf{Particle interaction graph.}
Following Graph Network-based Simulator (GNS)~\citep{sanchez2020learning}, we represent interactions by $\mathcal{G}_{t}=(\mathcal{V}_{t},\mathcal{E}_{t})$, whose nodes comprise object particles and controls. A spatial graph defines incoming object neighbors as $\mathcal{N}_{t}(i)=\{j\neq i:\|\mathbf{x}_{j}^{t}-\mathbf{x}_{i}^{t}\|_{2}<r\}$, while control--object edges convey manipulation. Node features encode motion history and attributes; directed edge features encode relative positions and distances. Their latent representations at processor layer $l$ are denoted by $\mathbf{h}_{i}^{l}$ and $\mathbf{e}_{ji}^{l}$.

\textbf{Graph-based dynamics prediction.}
An encoder--processor--decoder network performs $L$ message-passing layers, updating edge representations and aggregating incoming messages at each node. The decoder predicts a discrete acceleration for each object particle:
\begin{equation}
    \hat{\mathbf{a}}_{i}^{t}=\phi_{\mathrm{dec}}(\mathbf{h}_{i}^{L}), \qquad \hat{\mathbf{x}}_{i}^{t+1}=2\mathbf{x}_{i}^{t}-\mathbf{x}_{i}^{t-1}+\hat{\mathbf{a}}_{i}^{t}.
    \label{eq:particle_update}
\end{equation}
Here, $\hat{\mathbf{a}}_{i}^{t}$ denotes the change in per-frame displacement, with the fixed time step absorbed.

\section{Methodology}
\label{sec:method}

\begin{figure}[t]
    \centering
    \includegraphics[width=\linewidth]{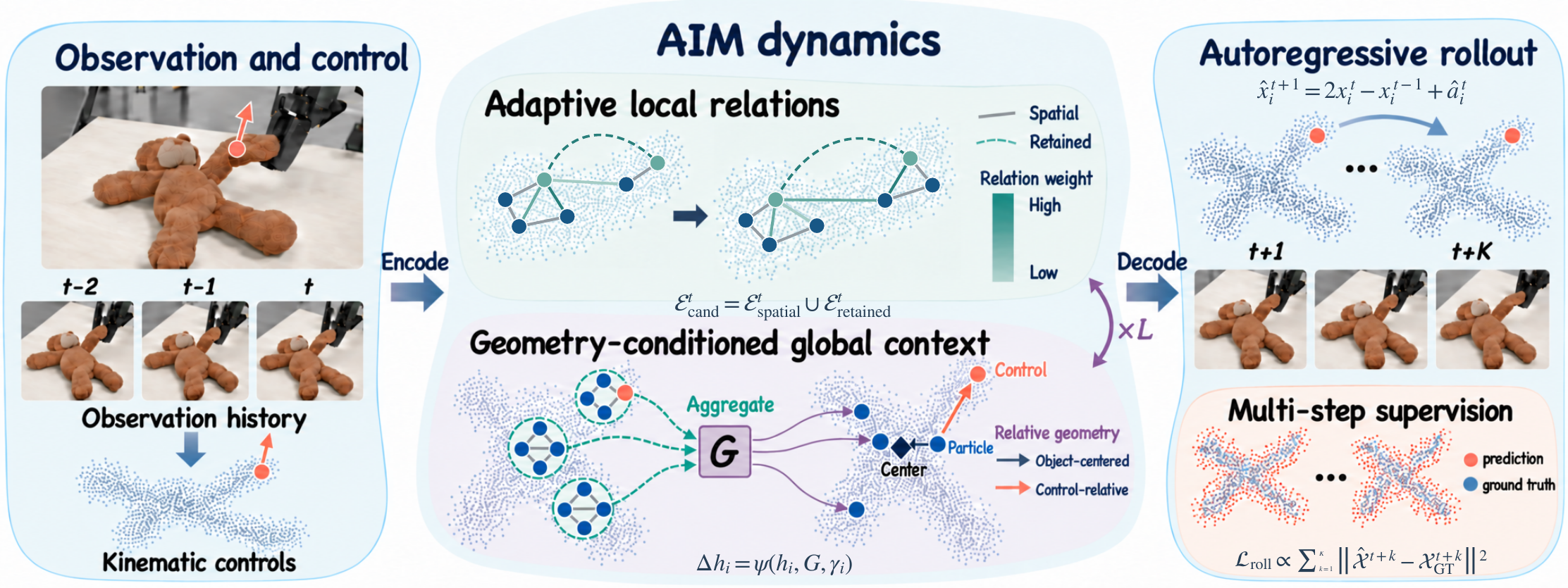}
    \vspace{-14pt}
    \caption{Overview of \themodel. Adaptive particle relations and geometry-conditioned global communication jointly predict object dynamics under kinematic controls.}
    \label{fig:aim_framework}
    \vspace{-12pt}
\end{figure}

\themodel models deformable dynamics by adapting particle interactions to the evolving deformation and coordinating local responses through object-level context. Conditioned on observation history and kinematic controls, it learns to predict object motion through multi-step autoregressive training. Figure~\ref{fig:aim_framework} provides an overview of the framework.

\subsection{Adaptive Interaction Representation}
\label{sec:adaptive_graph}

We represent local interactions through a history-conditioned graph. Current spatial neighborhoods provide candidate connections, while retained relations preserve communication across changing configurations. Each candidate receives a learned confidence and relation embedding, determining both its contribution to message passing and the information carried by its edge.

\textbf{Candidate relations.}
For object particles, we construct the candidate set as
\begin{equation}
    \mathcal{E}_{t}^{\mathrm{cand}}=\mathcal{E}_{t}^{\mathrm{spatial}}\cup\mathcal{E}_{t-1}^{\mathrm{retained}},
    \label{eq:candidate_graph}
\end{equation}
where $\mathcal{E}_{t}^{\mathrm{spatial}}$ contains current spatial neighbors and $\mathcal{E}_{t-1}^{\mathrm{retained}}$ contains confident relations retained from the preceding step. Retained pairs are rescored using the current state, allowing previously informative connections to remain candidates even when stretching separates their endpoints beyond the spatial radius.

Kinematic controls connect to object particles through bidirectional proximity edges whenever $\|\mathbf{x}_{i}^{t}-\mathbf{c}_{j}^{t}\|_{2}<\rho_j+\delta$, where $\rho_j$ is the control interaction radius and $\delta$ is a spatial margin. These edges have unit message weights and are reconstructed at each step without relation retention. Neighborhood radii, neighbor budgets, and retention settings are provided in the Appendix~\ref{app:graph_construction}.

\textbf{History-conditioned relation inference.}
We encode each object particle into a descriptor $\mathbf{z}_{i}^{t}=\phi_{\mathrm{hist}}([\mathbf{H}_{i}^{t},\mathbf{G}_{i}^{t},\mathbf{C}_{i}^{t}])$. Here, $\mathbf{H}_{i}^{t}$ describes recent displacements relative to the object's mean motion, $\mathbf{G}_{i}^{t}$ summarizes neighborhood distances and degree, and $\mathbf{C}_{i}^{t}$ contains the nearest control's relative position and attributes. For each candidate pair $(i,j)$, we measure descriptor compatibility through $A$ learned similarity channels:
\begin{equation}
    s_{ij}^{t,(a)}=\frac{\langle\mathbf{z}_{i}^{t}\odot\mathbf{w}_{a},\,\mathbf{z}_{j}^{t}\odot\mathbf{w}_{a}\rangle}{\|\mathbf{z}_{i}^{t}\odot\mathbf{w}_{a}\|_{2}\|\mathbf{z}_{j}^{t}\odot\mathbf{w}_{a}\|_{2}},\qquad a=1,\ldots,A,
    \label{eq:multihead_cos}
\end{equation}
where $\mathbf{w}_{a}$ is a learnable feature-wise weighting vector and $\odot$ denotes element-wise multiplication. We concatenate these similarities into $\mathbf{s}_{ij}^{t}$ and infer a relation embedding and confidence:
\begin{equation}
    \begin{aligned}
        \mathbf{q}_{ij}^{t}&=\phi_{\mathrm{pair}}\!\left([\mathbf{s}_{ij}^{t},|\mathbf{z}_{i}^{t}-\mathbf{z}_{j}^{t}|,\mathbf{z}_{i}^{t}\odot\mathbf{z}_{j}^{t},\boldsymbol{\psi}_{ij}^{t}]\right),\\
        p_{ij}^{t}&=\sigma\!\left(\phi_{\mathrm{score}}(\mathbf{q}_{ij}^{t})\right),\qquad
        w_{ij}^{t}=p_{ij}^{t}\mathbb{I}[p_{ij}^{t}\geq\tau].
    \end{aligned}
    \label{eq:relation_inference}
\end{equation}
The auxiliary descriptor $\boldsymbol{\psi}_{ij}^{t}=[\boldsymbol{\psi}_{ij}^{\mathrm{mot}},\boldsymbol{\psi}_{ij}^{\mathrm{geo}}]$ captures pair separation, relative-motion statistics, and local geometric compatibility; its construction is detailed in Appendix~\ref{app:relation_encoding}. We use $\tau=0.5$ to obtain sparse weighted connections, with a soft-weight warm-up followed by straight-through training as detailed in Appendix~\ref{app:training_details}. Both directed edges of an object pair share $\mathbf{q}_{ij}^{t}$ and $w_{ij}^{t}$. Relations are inferred once per prediction step and reused across its $L$ processor layers.

\textbf{Relation-aware graph encoding.}
Node features comprise normalized motion history, spatial boundary features, and node-role embeddings; control nodes additionally encode prescribed displacements and control attributes. A node encoder maps these features to $\mathbf{h}_{i}^{0}$. For a directed edge $j\!\rightarrow\!i$, we combine its geometric encoding with the inferred relation:
\begin{equation}
    \mathbf{e}_{ji}^{0}=\phi_{\mathrm{enc}}^{e}(\mathbf{f}_{ji}^{e,t})+\mathbf{b}_{\kappa(ji)}+\mathbf{W}_{q}\mathbf{q}_{ij}^{t},
    \label{eq:relation_encoding}
\end{equation}
where $\mathbf{f}_{ji}^{e,t}$ contains relative positions, distances, and their temporal differences, $\mathbf{b}_{\kappa(ji)}$ is a learned embedding identifying the object--object or control--object channel, and $\mathbf{W}_{q}$ projects the relation embedding into the edge latent space. We set $\mathbf{q}_{ij}^{t}=\mathbf{0}$ for control edges. The encoded graph therefore carries both geometric features and inferred interaction information, with $w_{ij}^{t}$ regulating the subsequent message aggregation.

\subsection{Local--Global Dynamics Propagation}
\label{sec:local_global}

We couple relation-weighted local propagation with an object-level latent, following the node--edge--global formulation of graph networks~\citep{battaglia2018relational}. The global latent aggregates object and control states and redistributes their shared context according to particle geometry, allowing distant regions to exchange information beyond local graph neighborhoods. We omit the physical time index on node and edge features below.

\textbf{Global state initialization.}
At each prediction step, we initialize a global token from separately pooled object and control features and an explicit motion summary. Let $\mathbf{H}^{l}$ stack the node features, and let $\mathcal{P}_{\mathrm{obj}}$ and $\mathcal{P}_{\mathrm{ctrl}}$ denote mean pooling over object and control nodes, respectively. The motion summary combines the mean object displacement with its first-order spatial moment:
\begin{equation}
    \boldsymbol{\xi}_{t}=\left[\frac{1}{N}\sum_{i=1}^{N}\tilde{\mathbf{v}}_{i}^{t},\,\frac{1}{N}\sum_{i=1}^{N}\frac{\mathbf{x}_{i}^{t}-\bar{\mathbf{x}}^{t}}{r}\times\tilde{\mathbf{v}}_{i}^{t}\right],
    \label{eq:motion_moment}
\end{equation}
where $\tilde{\mathbf{v}}_{i}^{t}$ is the normalized displacement from the previous frame, $\bar{\mathbf{x}}^{t}$ is the object centroid, and $r$ is the interaction scale. These statistics describe collective translation and the spatial organization of motion around the object center. We initialize the token as $\mathbf{g}_{t}^{0}=\phi_{\mathrm{init}}([\mathcal{P}_{\mathrm{obj}}(\mathbf{H}^{0}),\mathcal{P}_{\mathrm{ctrl}}(\mathbf{H}^{0}),\boldsymbol{\xi}_{t}])$.

\textbf{Relation-weighted local propagation.}
At processor layer $l$, each directed edge computes an increment $\Delta\mathbf{e}_{ji}^{l}=\phi_{e}^{l}([\mathbf{e}_{ji}^{l},\mathbf{h}_{j}^{l},\mathbf{h}_{i}^{l}])$ and updates its latent through $\mathbf{e}_{ji}^{l+1}=\mathbf{e}_{ji}^{l}+\Delta\mathbf{e}_{ji}^{l}$. Incoming increments are weighted by the inferred relations to update each node:
\begin{equation}
    \mathbf{m}_{i}^{l}=\sum_{j\rightarrow i}w_{ji}^{t}\Delta\mathbf{e}_{ji}^{l},\qquad
    \tilde{\mathbf{h}}_{i}^{l+1}=\mathbf{h}_{i}^{l}+\phi_{v}^{l}\!\left([\mathbf{h}_{i}^{l},\mathbf{m}_{i}^{l}]\right).
    \label{eq:adaptive_message}
\end{equation}
Object--object messages use the learned weights from Sec.~\ref{sec:adaptive_graph}, while control connections retain unit weights. The relation embedding determines message content through the edge latent, and the relation weight regulates its contribution to the receiving node.

\textbf{Geometry-conditioned global updates.}
After local propagation, we refine the global token using the updated node features:
\begin{equation}
    \mathbf{g}_{t}^{l+1}=\mathbf{g}_{t}^{l}+\phi_{g}^{l}\!\left([\mathbf{g}_{t}^{l},\mathcal{P}_{\mathrm{obj}}(\tilde{\mathbf{H}}^{l+1}),\mathcal{P}_{\mathrm{ctrl}}(\tilde{\mathbf{H}}^{l+1})]\right).
    \label{eq:global_update}
\end{equation}
Separate pooling distinguishes the evolving object state from the prescribed manipulation. We localize the shared context using a descriptor for each object particle, $\boldsymbol{\gamma}_{i}^{t}=[(\mathbf{x}_{i}^{t}-\bar{\mathbf{x}}^{t})/r,\,(\mathbf{c}_{n(i)}^{t}-\mathbf{x}_{i}^{t})/r]$, where $n(i)$ indexes the nearest control point. The descriptor captures the particle's position relative to the object center and its nearest control to condition the global residual:
\begin{equation}
    \mathbf{h}_{i}^{l+1}=\tilde{\mathbf{h}}_{i}^{l+1}+\phi_{p}^{l}\!\left([\tilde{\mathbf{h}}_{i}^{l+1},\mathbf{g}_{t}^{l+1},\boldsymbol{\gamma}_{i}^{t}]\right),\qquad i=1,\ldots,N.
    \label{eq:geometry_broadcast}
\end{equation}
Control-node features retain their local updates. The completed node states enter the next processor, so global information also conditions subsequent local messages. Each layer has independent parameters, with its update functions shared across nodes or edges. Geometry descriptors remain fixed within a prediction step, whereas the token evolves across processor layers and is initialized anew at the next physical time step.

\subsection{State Prediction and Training}
\label{sec:prediction_training}

The final particle representations are decoded into motion increments and recursively integrated to generate a rollout. We train the complete framework on these predicted trajectories, supervising both immediate state accuracy and errors propagated through subsequent steps.

\textbf{Particle state prediction.}
For each object particle, the decoder predicts a normalized discrete acceleration. We restore its coordinate scale as $\hat{\mathbf{a}}_{i}^{t}=\boldsymbol{\mu}_{a}+\boldsymbol{\sigma}_{a}\odot\phi_{\mathrm{dec}}(\mathbf{h}_{i}^{L})$, where $\boldsymbol{\mu}_{a}$ and $\boldsymbol{\sigma}_{a}$ are fixed statistics computed from the training data, and integrate it using Eq.~\ref{eq:particle_update}. Predicted positions are appended to the observation history, and the updated relation record is carried into the next step. Control positions continue to follow the prescribed trajectory.

\textbf{Autoregressive supervision.}
Training rollouts start from an observed history and proceed for $K$ steps using the model's own predicted object states. Ground-truth positions provide supervision without resetting intermediate predictions. For a batch of $B$ trajectories, we minimize
\begin{equation}
    \mathcal{L}_{\mathrm{roll}}=\frac{\lambda}{BK}\sum_{b=1}^{B}\sum_{k=1}^{K}\frac{1}{3|\mathcal{V}_{bk}|}\sum_{i\in\mathcal{V}_{bk}}\|\hat{\mathbf{x}}_{bki}-\mathbf{x}_{bki}^{\mathrm{GT}}\|_{2}^{2},
    \label{eq:rollout_loss}
\end{equation}
where $\mathcal{V}_{bk}$ is the set of particles with valid supervision for sample $b$ at rollout step $k$, and $\lambda$ scales the objective. Averaging over valid coordinates gives each sample and rollout step equal weight. Gradients propagate through the complete rollout without detaching predicted positions, allowing later errors to influence earlier state updates. The relation inference and dynamics components are jointly optimized without auxiliary supervision for edges or physical parameters. Details of autoregressive state updates, gradient flow, and optimization are provided in Appendix~\ref{app:training_details}.

\section{Experimental Settings}

\subsection{Datasets and Tasks}
We evaluate \themodel on two public real-world deformable-object benchmarks. \textbf{PhysTwin}~\citep{jiang2025phystwin} provides multi-view human--object interaction sequences and reconstructed appearance assets; we use its 22 main scenes and additional clothing sequences for resimulation, future prediction, and generalization across scenes, objects, and actions. \textbf{PGND}~\citep{zhang2025particle} covers robot interactions with six object categories: cloth, rope, plush, box, paper bag, and bread. We evaluate prediction under held-out interaction trajectories and cross-dataset transfer from PGND to PhysTwin. Together, these experiments assess within-domain predictive fidelity and cross-condition generalization, including transfer from robot to human manipulation. Dataset details, splits, and task protocols are provided in Appendix~\ref{app:dataset}.

\subsection{Baselines}
We compare \themodel with three representative approaches: \emph{(i) PhysTwin}~\citep{jiang2025phystwin} identifies object-specific physical models from observations; \emph{(ii) PGND}~\citep{zhang2025particle} learns particle-based dynamics; and \emph{(iii) SoMA}~\citep{huang2026soma} models soft-body dynamics using Gaussian representations. We use matched observations, control trajectories, and test splits wherever supported by each method. Training and evaluation details are provided in Appendix~\ref{app:implementation}.

\subsection{Evaluation metrics.}
We evaluate both 3D dynamics accuracy and rendered appearance quality. 
For particle-space evaluation, we report MDE, Chamfer Distance (CD), EMD, and, when correspondence annotations are available, tracking error (Track). 
For image-space evaluation, we report IoU, Boundary F-score, PSNR, SSIM, and LPIPS~\citep{wang2004image,zhang2018unreasonable}. 
Metric definitions and dataset-specific evaluation protocols are provided in the Appendix~\ref{app:metric}.

\section{Results}
\subsection{Within-Domain Dynamics Prediction}
\label{sec:within_domain_prediction}

\begin{figure}[t] 
\centering
\includegraphics[width=1.0\linewidth]{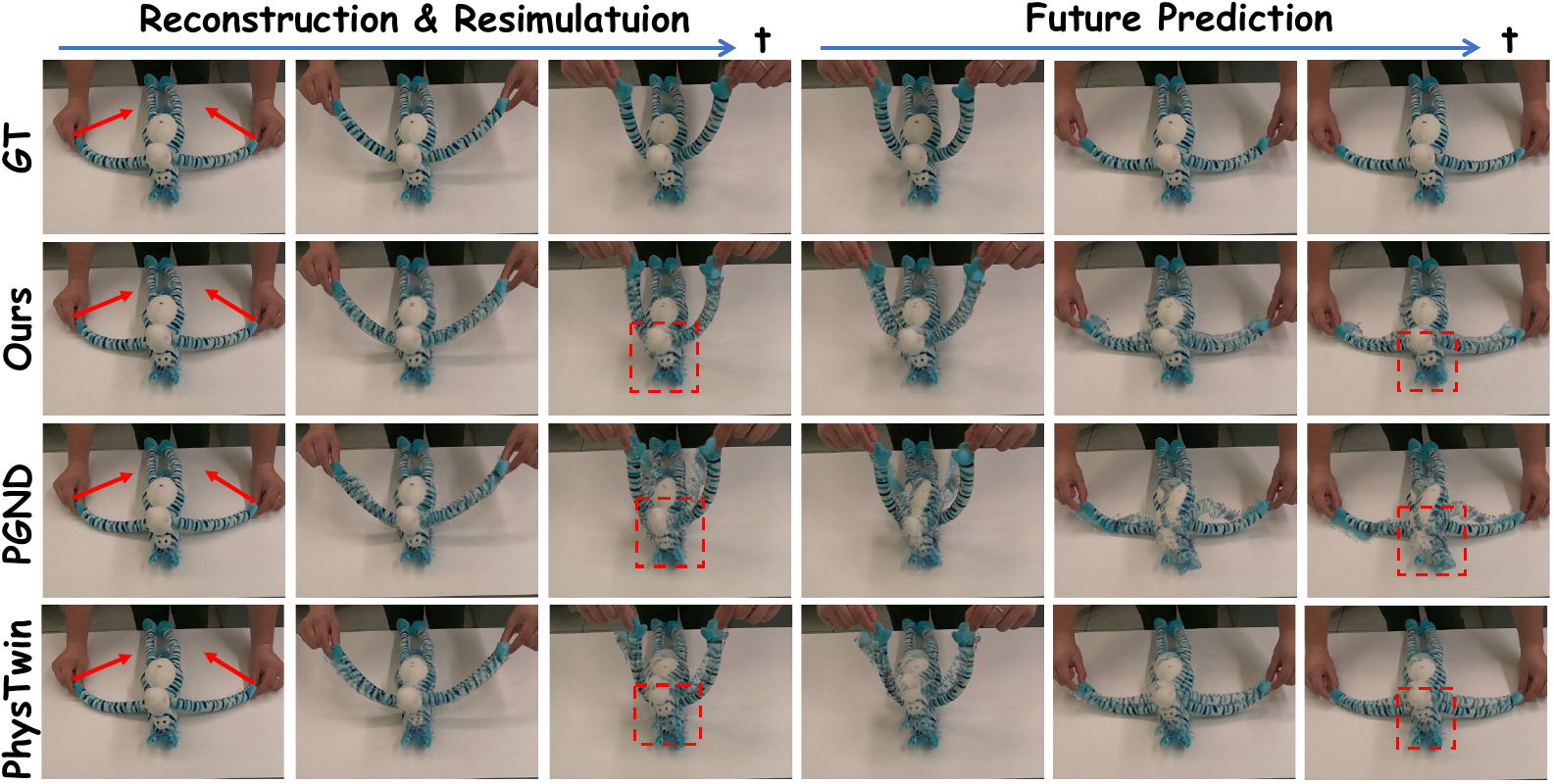}
\vspace{-14pt}
\caption{Qualitative Results on Reconstruction \& Resimulation and Future Prediction.}
\label{fig:phystwin_main}
\vspace{-12pt}
\end{figure}

We evaluate within-domain dynamics on PhysTwin through two tasks: reconstruction and resimulation reproduce observed interactions under recorded control trajectories, while future prediction extends the same autoregressive rollout beyond the observed interval without intermediate ground-truth resets. Table~\ref{tab:phystwin_main} reports quantitative results, and Fig.~\ref{fig:phystwin_main} presents qualitative comparisons.

\themodel achieves the best tracking accuracy and all four image-space metrics in both tasks. \emph{\textbf{(i)} Reproducing observed motion.} In reconstruction and resimulation, tracking error decreases by 30\% relative to PhysTwin, indicating more faithful motion of corresponding object points throughout deformation. Figure~\ref{fig:phystwin_main} (left) illustrates this advantage during lifting, with fewer local shape distortions in the highlighted central region. \emph{\textbf{(ii)} Predicting subsequent dynamics.} Beyond the observed interval, \themodel retains a 20\% tracking-error reduction, showing that the improvement extends to temporal extrapolation. The future configurations in Fig.~\ref{fig:phystwin_main} (right) more closely follow the ground truth as manipulation continues. Although PhysTwin maintains a slightly lower CD in both settings, the tracking and visual results support more accurate prediction of the evolving deformation and its projected appearance. Additional qualitative comparisons are provided in Appendix~\ref{app:phystwin_qualitative}.

\begin{table*}[t]
\centering
\caption{
Quantitative comparison on \textbf{reconstruction \& resimulation} and
\textbf{future prediction} on the PhysTwin benchmark.
Best results are shown in \textbf{bold}, and second-best results are \underline{underlined}.
}
\label{tab:phystwin_main}

\setlength{\tabcolsep}{4pt}
\renewcommand{\arraystretch}{1.05}

\resizebox{\linewidth}{!}{%
\begin{tabular}{lcccccc@{\hspace{8pt}}cccccc}
\toprule
& \multicolumn{6}{c}{\textbf{Reconstruction \& Resimulation}}
& \multicolumn{6}{c}{\textbf{Future Prediction}} \\
\cmidrule(lr){2-7}
\cmidrule(lr){8-13}

\textbf{Method}
& CD $\downarrow$
& Track $\downarrow$
& IoU $\uparrow$
& PSNR $\uparrow$
& SSIM $\uparrow$
& LPIPS $\downarrow$
& CD $\downarrow$
& Track $\downarrow$
& IoU $\uparrow$
& PSNR $\uparrow$
& SSIM $\uparrow$
& LPIPS $\downarrow$ \\
\midrule

PhysTwin
& \textbf{0.005}
& 0.010
& 84.4
& 28.636
& 0.949
& 0.030
& \textbf{0.012}
& 0.025
& 68.2
& 25.321
& 0.942
& 0.056 \\

SoMA
& 0.018
& 0.058
& 67.0
& 25.007
& 0.917
& 0.100
& 0.041
& 0.148
& 39.5
& 21.776
& 0.865
& 0.136 \\

PGND
& 0.009
& 0.013
& 82.1
& 28.228
& 0.949
& 0.033
& 0.024
& 0.044
& 60.3
& 24.426
& 0.938
& 0.069 \\

\rowcolor{oursbg}
\themodel
& \underline{0.006}
& \textbf{0.007}
& \textbf{87.0}
& \textbf{29.473}
& \textbf{0.952}
& \textbf{0.025}
& \underline{0.014}
& \textbf{0.020}
& \textbf{69.8}
& \textbf{25.916}
& \textbf{0.943}
& \textbf{0.055} \\

\bottomrule
\end{tabular}%
}
\vspace{-12pt}
\end{table*}

\subsection{Long-Horizon Rollout Stability}
\label{sec:long_horizon}

\begin{figure}[t] 
    \centering
    \includegraphics[width=1.0\linewidth]{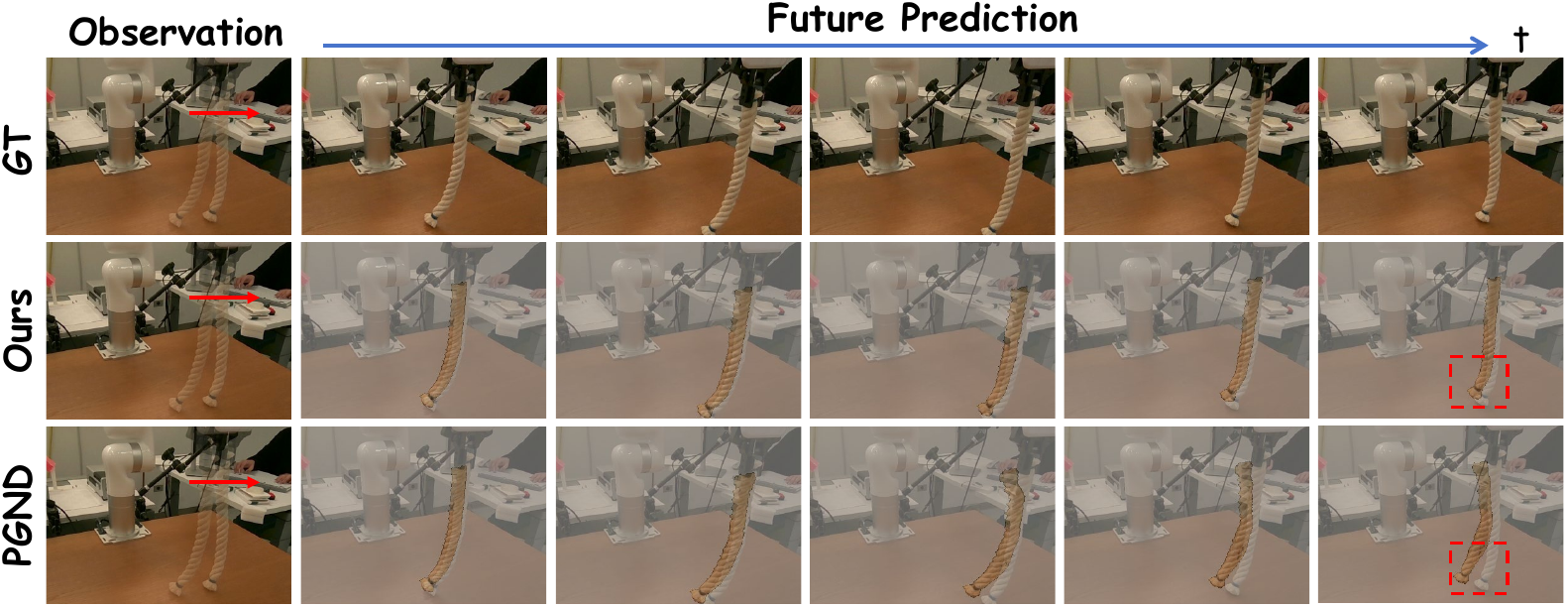}
    \vspace{-14pt}
    \caption{Qualitative Results on Held-out Interaction Prediction.}
    \label{fig:pgnd_rollout_inst_rope}
    \vspace{-12pt}
\end{figure}

\begin{figure}[t]
    \centering
    \includegraphics[width=\linewidth]{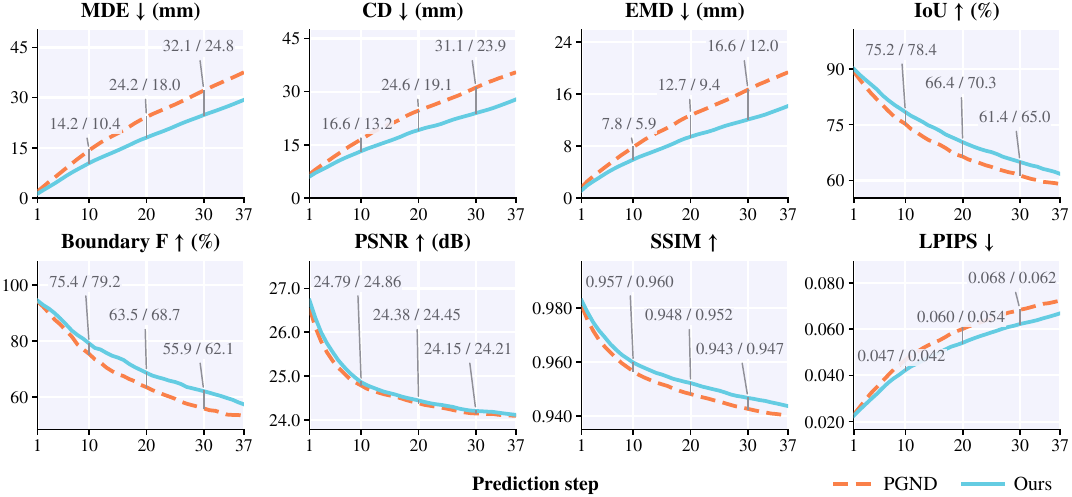}
    \vspace{-14pt}
    \caption{Long-Horizon Rollout Performance on Held-out Interaction Prediction.}
    \label{fig:pgnd_avg_metric}
    \vspace{-12pt}
\end{figure}

We evaluate 37-step autoregressive rollouts on held-out PGND interaction trajectories without intermediate ground-truth resets. Figure~\ref{fig:pgnd_avg_metric} reports metrics averaged equally across six object categories. Although both methods accumulate error over time, \themodel maintains more accurate motion predictions, with its absolute advantage increasing as the rollout progresses. At step 30, MDE is approximately 23\% lower than PGND, suggesting that the benefit persists through repeated prediction and integration. The rope sequence in Fig.~\ref{fig:pgnd_rollout_inst_rope} illustrates this difference: PGND develops increasingly visible drift near the free end, while \themodel remains more closely aligned with the observed configuration. This agreement between the temporal error trends and the visible endpoint drift supports improved long-horizon motion coherence under unseen control trajectories. Detailed per-category numerical results are provided in Appendix~\ref{app:pgnd_rollout}.

\begin{table*}[t]
\centering
\caption{Quantitative comparison on \textbf{cross-action transfer} and \textbf{cross-object transfer} on the PhysTwin benchmark. Best results are shown in \textbf{bold}. ``--'' indicates that transferring PhysTwin's object-specific spring-mass model requires additional adaptation beyond our evaluation protocol.}
\label{tab:phystwin_transfer_mean}
\setlength{\tabcolsep}{4.0pt}
\renewcommand{\arraystretch}{1.08}
\resizebox{\textwidth}{!}{%
\begin{tabular}{lcccccccccccc}
\toprule
& \multicolumn{6}{c}{\textbf{Cross-Action Transfer}}
& \multicolumn{6}{c}{\textbf{Cross-Object Transfer}} \\
\cmidrule(lr){2-7}\cmidrule(lr){8-13}
\textbf{Method}
& CD $\downarrow$ & Track $\downarrow$ & IoU $\uparrow$ & PSNR $\uparrow$ & SSIM $\uparrow$ & LPIPS $\downarrow$
& CD $\downarrow$ & Track $\downarrow$ & IoU $\uparrow$ & PSNR $\uparrow$ & SSIM $\uparrow$ & LPIPS $\downarrow$ \\
\midrule
PhysTwin
& 0.035 & 0.053 & 53.1 & 21.679 & 0.900 & 0.110
& -- & -- & -- & -- & -- & -- \\
PGND
& 0.226 & 0.274 & 58.4 & 23.763 & 0.902 & 0.089
& 0.026 & 0.038 & 72.2 & 23.023 & 0.934 & 0.086 \\
\rowcolor{oursbg}
\themodel
& \textbf{0.020} & \textbf{0.045} & \textbf{70.6} & \textbf{24.430} & \textbf{0.904} & \textbf{0.073}
& \textbf{0.018} & \textbf{0.023} & \textbf{79.8} & \textbf{24.506} & \textbf{0.942} & \textbf{0.073} \\
\bottomrule
\end{tabular}%
}
\vspace{-12pt}
\end{table*}

\subsection{Generalization Across Interaction Conditions}
\label{sec:generalization}

We evaluate generalization under three transfer settings, keeping the learned dynamics frozen during target evaluation. On PhysTwin, single-scene models are transferred to different actions or object instances, while shared category-level models are evaluated on held-out cloth scenes and folding interactions excluded from lifting-only training. We further assess zero-shot cross-dataset transfer from PGND robot interactions to human-manipulated PhysTwin sequences after coordinate and temporal alignment. Together, these settings test transfer across objects, actions, scenes, and manipulation domains without target-specific dynamics training.

\textbf{Single-Scene Generalization.}
Table~\ref{tab:phystwin_transfer_mean} evaluates transfer from individually trained scene models to new actions or object instances without target-specific training. \emph{\textbf{(i)} Generalization across actions.} Across the three action-transfer pairs, \themodel achieves the best average performance on all reported metrics, reducing CD by approximately 44\% relative to PhysTwin. The accompanying improvements in projected shape and appearance indicate better preservation of object geometry under changed manipulation patterns, including lifting-to-stretching and stretching-to-lifting transfers. \emph{\textbf{(ii)} Generalization across objects.} Under matched action types, \themodel outperforms PGND across all metrics averaged over the two object-transfer pairs. The approximately 39\% reduction in tracking error highlights its ability to predict motion on different object instances, with consistent visual improvements supporting transfer beyond the source geometry. Detailed per-pair quantitative comparisons and qualitative analyses are provided in Appendices~\ref{app:phystwin_cross_action}.

\textbf{Category-Level Generalization.}
Table~\ref{tab:phystwin_generalization} evaluates shared models on held-out cloth scenes and an unseen action type. \emph{\textbf{(i)} Generalization to held-out scenes.} \themodel reduces CD by approximately 48\% relative to PGND, accompanied by consistent improvements in rendered appearance. This agreement supports more faithful deformation prediction across the evaluated variations in object geometry and scene appearance. \emph{\textbf{(ii)} Generalization to unseen actions.} A shared model trained on lifting also transfers to folding, reducing CD by approximately 41\% relative to PGND. The clearest benefit is in geometric response and silhouette alignment, while perceptual improvements are more limited. These results extend the evidence for transfer beyond individual scene pairs to an action type excluded from shared-model training. Detailed per-scene quantitative results and qualitative comparisons for scene and action generalization are provided in Appendices~\ref{app:phystwin_cross_object}.

\begin{table*}[t]
\centering
\caption{Generalization to \textbf{held-out cloth scenes} and \textbf{unseen action types}. Best results are shown in \textbf{bold}. ``--'' indicates unavailable ground-truth 3D tracking annotations.}
\label{tab:phystwin_generalization}
\setlength{\tabcolsep}{4.0pt}
\renewcommand{\arraystretch}{1.08}
\resizebox{\textwidth}{!}{%
\begin{tabular}{lcccccccccccc}
\toprule
& \multicolumn{6}{c}{\textbf{Held-Out Scene Generalization}}
& \multicolumn{6}{c}{\textbf{Held-Out Action Generalization: Lift $\rightarrow$ Fold}} \\
\cmidrule(lr){2-7}\cmidrule(lr){8-13}
\textbf{Method}
& CD $\downarrow$ & Track $\downarrow$ & IoU $\uparrow$ & PSNR $\uparrow$ & SSIM $\uparrow$ & LPIPS $\downarrow$
& CD $\downarrow$ & Track $\downarrow$ & IoU $\uparrow$ & PSNR $\uparrow$ & SSIM $\uparrow$ & LPIPS $\downarrow$ \\
\midrule
PGND
& 0.023 & -- & 84.3 & 24.118 & 0.923 & 0.090
& 0.032 & -- & 69.1 & 22.037 & \textbf{0.900} & 0.109 \\
\rowcolor{oursbg}
\themodel
& \textbf{0.013} & -- & \textbf{87.1} & \textbf{24.869} & \textbf{0.929} & \textbf{0.074}
& \textbf{0.019} & -- & \textbf{71.3} & \textbf{22.982} & 0.896 & \textbf{0.108} \\
\bottomrule
\end{tabular}%
}
\vspace{-12pt}
\end{table*}

\begin{wraptable}{l}{0.45\linewidth}
\vspace{-5mm}
\centering
\setlength{\abovecaptionskip}{0pt}
\setlength{\belowcaptionskip}{3pt}
\caption{Zero-shot transfer from PGND to PhysTwin. Best results are shown in \textbf{bold}.}
\label{tab:cross_dataset_transfer}
\setlength{\tabcolsep}{4pt}
\renewcommand{\arraystretch}{1.08}
\resizebox{\linewidth}{!}{%
\begin{tabular}{lcccccc}
\toprule
\textbf{Method} & CD $\downarrow$ & Track $\downarrow$ & IoU $\uparrow$ & PSNR $\uparrow$ & SSIM $\uparrow$ & LPIPS $\downarrow$ \\
\midrule
PGND & 0.019 & 0.033 & 60.28 & 24.835 & \textbf{0.942} & 0.070 \\
\rowcolor{oursbg}
\themodel & \textbf{0.016} & \textbf{0.023} & \textbf{65.21} & \textbf{25.438} & 0.941 & \textbf{0.069} \\
\bottomrule
\end{tabular}%
}
\end{wraptable}

\textbf{Zero-Shot Cross-Dataset Transfer.}
We transfer models trained on PGND robot interactions to human-manipulated PhysTwin sequences, keeping all dynamics parameters frozen after coordinate and temporal alignment. Averaged across three object categories, \themodel reduces tracking error by approximately 30\% relative to PGND (Table~\ref{tab:cross_dataset_transfer}), indicating more accurate motion prediction despite the change in manipulation domain. Improved object overlap further supports closer alignment with the observed deformation, while gains in perceptual quality are modest. These results show that the learned dynamics remain applicable across robot and human manipulation without target-domain dynamics fitting. Detailed quantitative results for the individual scenes summarized in Table~\ref{tab:cross_dataset_transfer}, together with the corresponding analyses, are provided in Appendix~\ref{app:cross_dataset_transfer}.

\begin{figure}[!htbp] 
    \centering
    \includegraphics[width=1.0\linewidth]{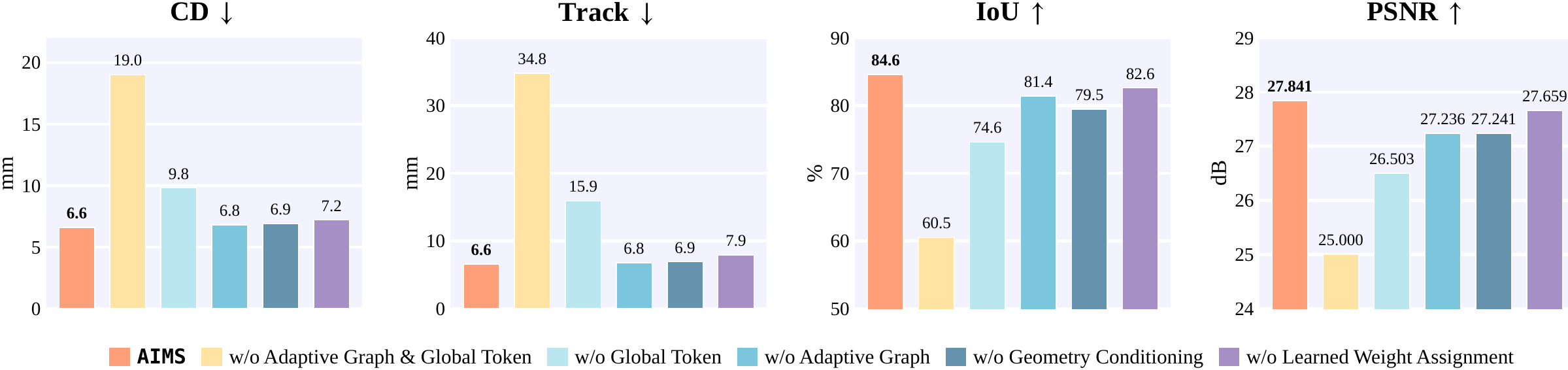}
    \vspace{-14pt}
    \caption{Ablation study on double\_stretch\_zebra}
    \label{fig:ablation}
    \vspace{-12pt}
\end{figure}

\subsection{Ablation Analysis}

We conduct single-component ablations on the double\_stretch\_zebra scene, following a one-change-at-a-time protocol. Figure~\ref{fig:ablation} shows that the full model consistently outperforms the ablated variants. The largest degradation follows the removal of object-level context, suggesting that local propagation alone is insufficient to coordinate particle motion. Adaptive relations provide complementary gains by adjusting communication to the evolving deformation, while geometric conditioning helps translate shared context into spatially appropriate responses. Together, these findings support combining adaptive local interactions with geometry-conditioned global communication. Variant definitions, checkpoint selection, and detailed results are provided in Appendix~\ref{app:ablation_details}.

\section{Conclusion}
In this work, we present \themodel, an adaptive interaction modeling framework for real-to-sim soft-body simulation that couples deformation-dependent local relations with geometry-conditioned global context. Evaluations on PhysTwin and PGND demonstrate improved motion tracking and visual fidelity, reduced error accumulation over long rollouts, and transfer across actions, object instances, and manipulation settings. These results highlight the value of jointly modeling evolving interactions and object-level context for accurate and transferable deformable dynamics. We hope this work supports future research on interactive digital twins and model-based soft-body manipulation.

\section*{AI Use Statement}
We used OpenAI Codex to assist to check grammar with \LaTeX{} formatting. It was not used to generate research data, design or run experiments, implement the reported method, formulate mathematical claims, assist with proofs, or compute the reported results. All AI-assisted edits were checked against the source manuscript and reported tables. The authors take full responsibility for the final text, claims, and artifacts

\section*{Ethics Statement}
This work studies real-to-sim simulation of deformable objects using existing public datasets. We do not collect new human-subject data. Images depicting human manipulation are used to study object responses to external interactions, rather than to infer personal attributes or behavior. Our experiments concern prediction and simulation; predictive accuracy alone does not establish the safety of deployment on physical robots. Any downstream deployment requires appropriate validation of contact behavior, failure modes, and operational safety under the intended interaction conditions.

\section*{Reproducibility Statement}
We provide detailed methodological descriptions, experimental protocols, and implementation settings to facilitate reproducibility. Section~\ref{sec:method} describes the proposed framework, with graph construction and relation encoding detailed in Appendices~\ref{app:graph_construction} and~\ref{app:relation_encoding}. Appendix~\ref{app:training_details} explains autoregressive supervision and gradient handling, while Appendix~\ref{app:implementation} specifies network configurations and training and evaluation settings. Dataset splits and task protocols are documented in Appendix~\ref{app:dataset}, and metric definitions are provided in Appendix~\ref{app:metric}. Appendix~\ref{app:ablation_details} describes the ablation variants and checkpoint selection. Code, configuration files, and instructions for environment setup, training, and evaluation are available at \href{https://anonymous.4open.science/r/AIM-F887/}{https://anonymous.4open.science/r/AIM-F887/}

\bibliography{iclr2027_conference}
\bibliographystyle{iclr2027_conference}

\newpage
\appendix

\section{Additional Related Work}
\label{app:related_work}

To complement the main related work in Section~\ref{sec:related_work}, we discuss three interconnected areas: (i) physics-based simulation and visual deformable modeling; (ii) relational dynamics and global communication; and (iii) autoregressive prediction and generalization across interaction conditions.

\textbf{Physics-Based and Visual Deformable Modeling.}
Finite element methods, material point methods, and position-based dynamics provide established discretizations for deformable simulation~\cite{belytschko2014nonlinear,stomakhin2013material,muller2007position}, whose application to observed objects requires identifying geometry, material response, contact behavior, and boundary conditions. Differentiable simulators facilitate this process through parameter optimization and control, including general differentiable physical computation in DiffTaichi~\cite{hu2019difftaichi} and efficient differentiation through implicit soft-body simulation in DiffPD~\cite{du2021diffpd}. Visual reconstruction offers complementary information: Dynamic 3D Gaussians recover persistent trajectories and appearance~\cite{luiten2024dynamic}, while PhysGaussian, reconstruction-based physical models, and PhysTwin connect visual assets to continuum or spring--mass simulation~\cite{xie2024physgaussian,zhong2024reconstruction,jiang2025phystwin}. PhysDreamer further incorporates video-generation priors for material estimation~\cite{zhang2024physdreamer}. Recent hybrid approaches increase flexibility through neural spring properties in NeuSpring~\cite{xu2026neuspring} and learned residual dynamics in PGRD and DeformMaster~\cite{patel2026learning,li2026deformmaster}. These approaches retain an explicit physical backbone, whereas \themodel directly learns interaction-conditioned particle updates from recorded trajectories.

\textbf{Relational Dynamics and Global Communication.}
Learning which regions interact and how information propagates are closely related problems in deformable dynamics. Neural Relational Inference learns latent interactions jointly with dynamics~\cite{kipf2018neural}, Dynamic Neural Relational Inference models evolving relations~\cite{graber2020dynamic}, and Visible Connectivity Dynamics infers cloth connectivity from partial point clouds~\cite{lin2022learning}. These methods motivate distinguishing spatial proximity from motion-relevant dependencies. Beyond local relations, hierarchical communication extends the range of information exchange: DPI-Net models interactions across particle hierarchies~\cite{li2018learning}, MultiScale MeshGraphNets and BSMS-GNN introduce coarse graph levels~\cite{fortunato2022multiscale,cao2023efficient}, and HOOD applies hierarchical message passing to clothing dynamics~\cite{grigorev2023hood}. Other representations support broader coordination through particle--grid exchange in PGND~\cite{zhang2025particle}, hierarchical center-of-mass systems in GausSim~\cite{shao2025gaussim}, or global latent updates in graph networks~\cite{battaglia2018relational}. \themodel combines adaptive local relations with shared object-level context. Motion history and geometry determine communication weights over spatial and retained candidates, while geometry-conditioned global updates coordinate particle responses without explicit material-topology supervision.

\textbf{Autoregressive Prediction and Interaction Generalization.}
Reliable learned simulation must handle both recursively generated states and changes in external interaction. To improve rollout robustness, GNS trains with noise-corrupted histories~\cite{sanchez2020learning}, Message Passing Neural PDE Solvers introduces pushforward training on model-generated states~\cite{brandstetter2022message}, PGRD incorporates temporal context and multi-step residual training~\cite{patel2026learning}, and SoMA uses coarse-to-fine temporal learning~\cite{huang2026soma}. Action-conditioned prediction also supports manipulation: visual foresight evaluates candidate actions through video prediction~\cite{finn2017deep}, particle dynamics models support multiple physical materials~\cite{li2018learning}, and Visible Connectivity Dynamics addresses cloth manipulation under partial observations~\cite{lin2022learning}. GS-Dynamics predicts responses to unseen robot actions from Gaussian-derived trajectories~\cite{zhang2024dynamic}, while PGND and SoMA develop interaction interfaces for particle- and Gaussian-based simulation~\cite{zhang2025particle,huang2026soma}. These settings involve distinct generalization requirements: temporal extrapolation and held-out trajectories differ from changes in action type, object instance, or manipulation domain. PhysTwin transfers identified object models to new interactions~\cite{jiang2025phystwin}, whereas PGND also evaluates unseen instances within object categories~\cite{zhang2025particle}. Our evaluation separates within-domain prediction from cross-condition generalization, combining autoregressive rollout assessment with transfer across actions, objects, and manipulation settings, including zero-shot transfer from PGND robot interactions to human-manipulated PhysTwin sequences.

\section{Additional Architectural Details}
\label{app:additional_architectural_details}

We supplement the main formulation with details of candidate graph construction, relation descriptors, and autoregressive optimization. We retain the notation of the main text and distinguish the physical prediction step $t$ from the processor layer index $l$.

\subsection{Interaction Graph Construction}
\label{app:graph_construction}

\textbf{Spatial candidates.}
At each prediction step, we construct object--object candidates from the current particle positions within an expanded neighborhood of the base interaction scale $r$. To limit the candidate count in densely sampled regions, each particle first selects its nearest neighbors within this neighborhood, subject to a spatial budget. We then symmetrize the selections: a pair is included if either endpoint selects the other. Consequently, the budget applies before symmetrization and is not a strict bound on the final node degree. Self-connections are explicitly excluded, and each unordered pair is stored once before being expanded into two directed edges for message passing.

\textbf{Relation retention and renewal.}
The relation record associates previously evaluated object pairs with their confidence scores. At the next step, pairs exceeding the retention threshold are eligible for reuse, with a separate per-particle budget favoring higher-confidence relations. These selections are symmetrized and merged with the current spatial candidates, removing duplicates.

Retention preserves a pair's eligibility for communication, rather than fixing its weight or latent representation. Every candidate, including a retained pair, is re-encoded and rescored using the current history and geometry. A retained pair can therefore remain available after its endpoints leave the spatial neighborhood, but its message contribution can still be suppressed by the current score. Conversely, a discarded pair can re-enter through a subsequent spatial search. The retention criterion and the message gate serve different purposes: the former determines which historical pairs are reconsidered, whereas the latter determines their current contribution to propagation.

The record is initialized empty for each independent rollout. Particle identities remain consistent within the rollout, and records are maintained separately for different samples. This mechanism introduces temporal continuity without assuming a fixed initial topology or interpreting the inferred edges as permanent physical bonds.

\textbf{Control connections.}
Control--object edges are rebuilt from the current configuration using each control's interaction radius and a spatial margin. Each valid connection is inserted in both directions with unit message weight. These edges allow control features to influence nearby particles and object features to update the control representations used in global pooling. Their latent features participate in message passing, while their physical positions remain prescribed. Control edges are excluded from relation retention, so a previous proximity connection does not persist after the control moves away.

\subsection{Relation Features and Graph Encoding}
\label{app:relation_encoding}

\textbf{History-conditioned particle descriptors.}
The relation encoder receives complementary motion, neighborhood, and control information. Motion history consists of consecutive particle displacements after subtracting the mean object displacement at each historical transition and scaling by $r$. This emphasizes motion relative to the object's collective translation. Neighborhood statistics summarize the mean and standard deviation of current candidate-pair distances together with the logarithm of the candidate degree.

Control context contains the normalized offset from the nearest control to the particle, the control's available kinematic attributes, and a control-presence indicator. The attributes encode orientation, angular motion, interaction radius, and openness where available. This descriptor provides local interaction context; the prescribed control displacement is supplied separately to the dynamics node encoder. A shared multilayer perceptron maps the concatenated descriptor to $\mathbf{z}_i^t$.

\textbf{Pairwise motion evidence.}
For an unordered candidate pair, let
$\boldsymbol{\delta}_{ij}^{t}=(\mathbf{x}_i^t-\mathbf{x}_j^t)/r$,
$d_{ij}^{t}=\|\boldsymbol{\delta}_{ij}^{t}\|_2$, and
$\Delta\boldsymbol{\delta}_{ij}^{t}
=\boldsymbol{\delta}_{ij}^{t}-\boldsymbol{\delta}_{ij}^{t-1}$.
We decompose the relative displacement into its signed radial component
$a_{ij}^{t}
=\langle\Delta\boldsymbol{\delta}_{ij}^{t},
\widehat{\boldsymbol{\delta}}_{ij}^{t}\rangle$
and tangential magnitude
$b_{ij}^{t}
=\|\Delta\boldsymbol{\delta}_{ij}^{t}
-a_{ij}^{t}\widehat{\boldsymbol{\delta}}_{ij}^{t}\|_2$,
where the unit direction uses a stabilized denominator.
The motion descriptor is
\begin{equation}
\boldsymbol{\psi}_{ij}^{\mathrm{mot}}
=
\left[
d_{ij}^{t},\;
d_{ij}^{t}-d_{ij}^{t-1},\;
\|\Delta\boldsymbol{\delta}_{ij}^{t}\|_2,\;
a_{ij}^{t},\;
b_{ij}^{t},\;
\frac{\|\Delta\boldsymbol{\delta}_{ij}^{t}\|_2^2}
     {\|\Delta\boldsymbol{\delta}_{ij}^{t}\|_2^2+\epsilon_{\mathrm{mot}}^2},\;
\operatorname{Var}_{s\in\mathcal{T}_t}(d_{ij}^{s})
\right],
\end{equation}
where $\mathcal{T}_t$ indexes the available history and
$\epsilon_{\mathrm{mot}}$ is a motion noise floor.
These terms distinguish current separation, relative approach or separation, tangential motion, and variation in pair distance over time. The bounded motion-amplitude term indicates whether the observed relative motion is appreciable compared with the noise floor; it does not by itself imply that the pair is physically connected.

\textbf{Local geometric compatibility.}
We describe local geometry at two neighborhood scales. At each scale, the covariance of neighboring relative offsets provides a principal direction $\mathbf{a}_i$ and a reliability score $c_i$. Reliability is determined by the normalized gap between the largest two covariance eigenvalues and is suppressed when the neighborhood contains insufficient samples. Thus, a nearly isotropic neighborhood does not contribute an arbitrary dominant direction.

For a pair direction $\widehat{\boldsymbol{\delta}}_{ij}$, define
$u_i=\langle\mathbf{a}_i,\widehat{\boldsymbol{\delta}}_{ij}\rangle^2$
and $u_j$ analogously. The descriptor at neighborhood scale $s$ is
\begin{equation}
\boldsymbol{\psi}_{ij}^{\mathrm{geo},(s)}
=
c_i^{(s)}c_j^{(s)}
\left[
\langle\mathbf{a}_i^{(s)},\mathbf{a}_j^{(s)}\rangle^2,\;
\frac{u_i^{(s)}+u_j^{(s)}}{2},\;
\left|u_i^{(s)}-u_j^{(s)}\right|,\;
1
\right].
\end{equation}
Concatenating the descriptors from both scales yields
$\boldsymbol{\psi}_{ij}^{\mathrm{geo}}$.
Squared inner products remove the sign ambiguity of covariance eigenvectors. The resulting features summarize agreement between local directions and their alignment with the pair, while the confidence factors suppress unreliable directional evidence. They require neither mesh connectivity nor a predefined surface or centerline representation.

\textbf{Symmetric relations and directed edge states.}
The pair encoder combines the auxiliary motion and geometry descriptors with the learned similarity channels, the absolute difference of endpoint descriptors, and their element-wise product. These inputs are symmetric under exchanging the endpoints, producing a shared relation embedding and confidence for both edge directions.

Directed edge features retain information that the symmetric relation does not: the sender-to-receiver displacement, its magnitude, and their changes since the preceding frame. Object--object features are scaled by $r$, while control-edge features use a scale that also accounts for the control interaction extent. The encoded geometric features receive an edge-type embedding and, for object--object edges, a projected relation embedding. Thus, the confidence controls how strongly an edge contributes, while the relation embedding conditions the content of its message. Sharing a pair confidence does not force the two directed messages to be identical.

\textbf{Node encoding and normalization.}
The dynamics node encoder uses displacement histories standardized by fixed training-derived statistics, clipped distances to the workspace boundaries, and a learned object/control role embedding. Control nodes additionally receive their kinematic attributes and prescribed displacement; the corresponding control-specific entries are zero for object nodes. The relation descriptor's object-centered history and the node encoder's standardized history serve different roles: the former emphasizes relative deformation, while the latter retains collective motion information for state prediction.

Relation encoders and scoring functions use multilayer perceptrons with SiLU activations. The dynamics encoders and processor updates use ReLU multilayer perceptrons, with layer normalization on the latent encodings and designated update outputs. The final state decoder has an unnormalized linear output, whose coordinate scale is recovered using the fixed second-difference statistics. Normalization statistics remain fixed during evaluation.

\subsection{Autoregressive Training and Optimization}
\label{app:training_details}

\textbf{Training rollout.}
Each training sample provides an initial object history, a prescribed control sequence, and future object states for supervision. After each prediction, the predicted object positions and the next prescribed control positions are appended to the history, replacing its oldest frame. Future object observations are used only as targets. Candidate construction, relation scoring, and control proximity connections are recomputed from the resulting state at every step.

The relation record is carried between prediction steps, whereas node and edge encodings and the global token are initialized anew from the current inputs. Within a prediction step, the candidate graph and relation weights remain fixed across processor layers, while the latent node, edge, and global states evolve. This separates temporal relation retention from iterative latent propagation.

\textbf{Supervision and gradient flow.}
The rollout objective averages coordinate-wise squared position errors over valid supervised particles, then over samples and prediction steps. This prevents samples with more valid annotations from receiving disproportionate weight. Missing supervision affects the loss mask rather than replacing predicted states with observations.

Predicted positions are not detached between rollout steps, allowing later errors to influence earlier state updates through the differentiable prediction paths. Discrete operations, including neighborhood selection, pair lookup, and retention ranking, are treated as fixed decisions during backpropagation. Analytic motion statistics and covariance-based geometric descriptors are also computed without gradients through their extraction. The descriptor encoders, relation scores, message-passing functions, global updates, and decoder are nevertheless jointly optimized by the trajectory objective. No edge labels, auxiliary relation losses, or physical-parameter targets are required.

\textbf{Learning thresholded relations.}
Training begins with continuous confidence weights so that all candidate relations can receive a learning signal. In the subsequent phase, the forward pass uses the same thresholded weights as inference, while a straight-through estimator supplies gradients:
\begin{equation}
w_{ij}^{\mathrm{train}}
=
p_{ij}
+
\operatorname{sg}
\left(
p_{ij}\,\mathbf{1}\{p_{ij}\geq\tau\}-p_{ij}
\right),
\end{equation}
where $\operatorname{sg}$ denotes stop-gradient and the physical time index is omitted for clarity. The forward value is the thresholded confidence, while the surrogate derivative with respect to $p_{ij}$ is one.

During training, candidate messages remain in the computation even when their forward weights are zero, preserving this surrogate gradient to the scoring network. During inference, zero-weight edges can be removed before message passing. Retention is still a discrete selection based on previously computed confidence; the straight-through estimator applies to message weights, not to the candidate search or retention procedure.

\textbf{Optimization.} Where a rollout curriculum is used, the training horizon increases from shorter to longer unrollings while retaining the same autoregressive objective. We optimize the network with Adam and gradient-norm clipping. Processor activation checkpointing reduces memory consumption by recomputing intermediate activations during backward propagation, without detaching the predicted trajectory. Dataset-specific rollout schedules, relation-weight warm-up durations, and optimization settings are provided in Appendix~\ref{app:implementation}.

\section{Experimental Setup}
\label{app:experimental_setup}

\subsection{Datasets and Preprocessing}
\label{app:dataset}

\textbf{PhysTwin.}
PhysTwin~\citep{jiang2025phystwin} contains multi-view RGB-D recordings of human interactions with deformable objects, including cloth, ropes, plush toys, and packages. We use 22 interaction sequences and split each sequence temporally into 70\% for fitting and 30\% for future prediction. Reconstruction and resimulation are evaluated over the observed interval, while future prediction is evaluated over the remaining interval. The benchmark also provides manually annotated point trajectories for tracking evaluation. Additional cloth scenes are used for the category-level
generalization experiments described in the main text.

\textbf{PGND.} PGND~\citep{zhang2025particle} records bimanual robot interactions across six material categories: cloth, rope, plush, boxes, paper bags, and bread. We use the provided training and evaluation partitions to assess dynamics prediction under held-out interactions. Long-horizon evaluation follows autoregressive predictions over 37 steps.

\textbf{Data preprocessing.} We reuse the official preprocessing outputs, including calibrated object point clouds, tracked trajectories, interaction controls, and reconstructed Gaussian appearance. The underlying pipelines extract object regions from RGB-D observations and recover geometric and motion information through calibrated projection and point tracking. We organize these outputs into particle histories and prescribed control trajectories for dynamics learning. Robot grippers and tracked human interactions are represented through the same kinematic control-point interface.

\subsection{Implementation Details}
\label{app:implementation}

\textbf{Computing Infrastructure}
We implement \themodel in Python 3.10.12 using PyTorch 2.7.1 with CUDA 12.8. Experiments are conducted on a server equipped with eight NVIDIA A100-SXM4 GPUs, each with 40\,GB of memory. Each training run uses a single GPU, while independent experiments are distributed across available devices. We reuse the benchmark-specific rendering and evaluation implementations for visual comparisons.

\textbf{Network configuration.} \themodel uses a three-frame motion history and ten message-passing processors. Node, edge, and global latent states have width 128, and the backbone multilayer perceptrons use two hidden layers of width 128. The relation encoder produces 32-dimensional embeddings with four learned similarity channels. Processor parameters are independent across layers, while update functions are shared across particles and relations. The complete dynamics network contains approximately 3.17 million trainable parameters, excluding the Gaussian appearance representation and external preprocessing models.

\textbf{Interaction graph.} Spatial candidates are collected within 1.5 times the base interaction radius, with at most 64 spatial neighbors per particle before symmetrization. Up to 16 previous relations per particle are retained as candidates when their scores exceed 0.55. Candidate relations are rescored at each prediction step, with a message-weight threshold of 0.5 at inference. Control connections are reconstructed from the current configuration and use unit message weights. Further details are provided in Appendices~\ref{app:graph_construction} and~\ref{app:relation_encoding}.

\textbf{Training protocols.} We use the same dynamics architecture on both benchmarks, with dataset-specific temporal supervision and optimization schedules. Autoregressive state updates, supervision masks, and gradient handling are detailed in Appendix~\ref{app:training_details}; the configurations used for each benchmark are specified below.

\textbf{Optimization.} All models are trained from scratch using Adam with an initial learning rate of $10^{-4}$, momentum coefficients $(0.9,0.999)$, a numerical stabilizer of $10^{-8}$, and no weight decay. We use a batch size of two without gradient accumulation and clip the gradient norm to 1. No additional noise is injected into the input particle trajectories. Learning-rate schedules and temporal sampling follow the respective benchmark training configurations.

\textbf{PhysTwin training.} We preserve the original frame rate and train for 20,000 optimizer updates using the original PhysTwin training setup adopted in our experiments. PhysTwin provides relatively short interaction sequences, and its single-scene setting fits a model to one object under a specific recorded manipulation. Retaining this training protocol preserves the temporal supervision and sampling setup used for the corresponding reconstruction, resimulation, and future-prediction comparisons. Category-level models and ablation variants follow the same PhysTwin optimization configuration, with their respective training data and model components. Checkpoints are saved every 500 updates.

\textbf{PGND training.} For the six-category PGND experiments, we use fixed five-step autoregressive supervision to align with the PGND training protocol. Each rollout begins with an observed history and recursively reuses predicted object positions without intermediate ground-truth resets or gradient detachment. This exposes the model to the predicted-state inputs encountered during evaluation on held-out interaction trajectories. We train for 100,000 optimizer updates with the exponential learning-rate schedule used in the original runs and a sampling interval of 0.1\,s. The network configuration, batch size, optimizer coefficients, and gradient-clipping threshold are shared with the PhysTwin experiments. The loss sums the five per-step position errors with a coefficient of 100, equivalent to $\lambda=500$ in Eq.~\ref{eq:rollout_loss}.

\textbf{Transfer evaluation.} Cross-action and cross-object evaluations reuse the corresponding source model with frozen dynamics parameters. All targets associated with the same source share the same source checkpoint. Zero-shot cross-dataset evaluation similarly reuses the PGND-trained model after coordinate and temporal alignment, without target-domain dynamics fitting.

\subsection{Evaluation Protocol}
\label{app:metric}

\textbf{Geometric and motion metrics.} We evaluate object geometry using Chamfer Distance (CD) and motion accuracy using tracking error (Track) on PhysTwin and Mean Distance Error (MDE) on PGND. Track measures the Euclidean error of annotated points with correspondences fixed at initialization, whereas MDE measures displacement error between corresponding particles. We retain each benchmark's CD implementation: PhysTwin uses a one-sided nearest-neighbor distance from observed to predicted points with the $\ell_1$ norm, while PGND uses its bidirectional implementation. We additionally report Earth Mover's Distance (EMD) on PGND using the benchmark's matching implementation. Geometric errors are reported in meters or millimeters, as specified in each table or figure.

\textbf{Image-space metrics.} Intersection over Union (IoU) and boundary F-score assess object overlap and contour agreement. PSNR, SSIM, and LPIPS measure rendered-image fidelity. Higher values indicate better performance for IoU, boundary F-score, PSNR, and SSIM; lower values are preferable for geometric errors and LPIPS. Comparisons use matched viewpoints and evaluation frames, with consistent benchmark-specific masking and rendering settings.

\textbf{Rollout and aggregation.} During evaluation, predicted object states are recursively fed back into the simulator, while control trajectories remain prescribed. PhysTwin results are first averaged over valid frames within each sequence and then equally across evaluated sequences. Transfer results are equally averaged over the corresponding source--target pairs. For the PGND temporal curves, we average each prediction step over evaluation episodes within each material category and then equally across the six categories. Unavailable tracking annotations are excluded and marked as ``--'' rather than assigned numerical values.

\section{Additional Results}
\label{app:add_results}

\subsection{Qualitative Results on Within-Domain Dynamics Prediction}
\label{app:phystwin_qualitative}

\begin{figure}[t] 
\centering
\includegraphics[width=1.0\linewidth]{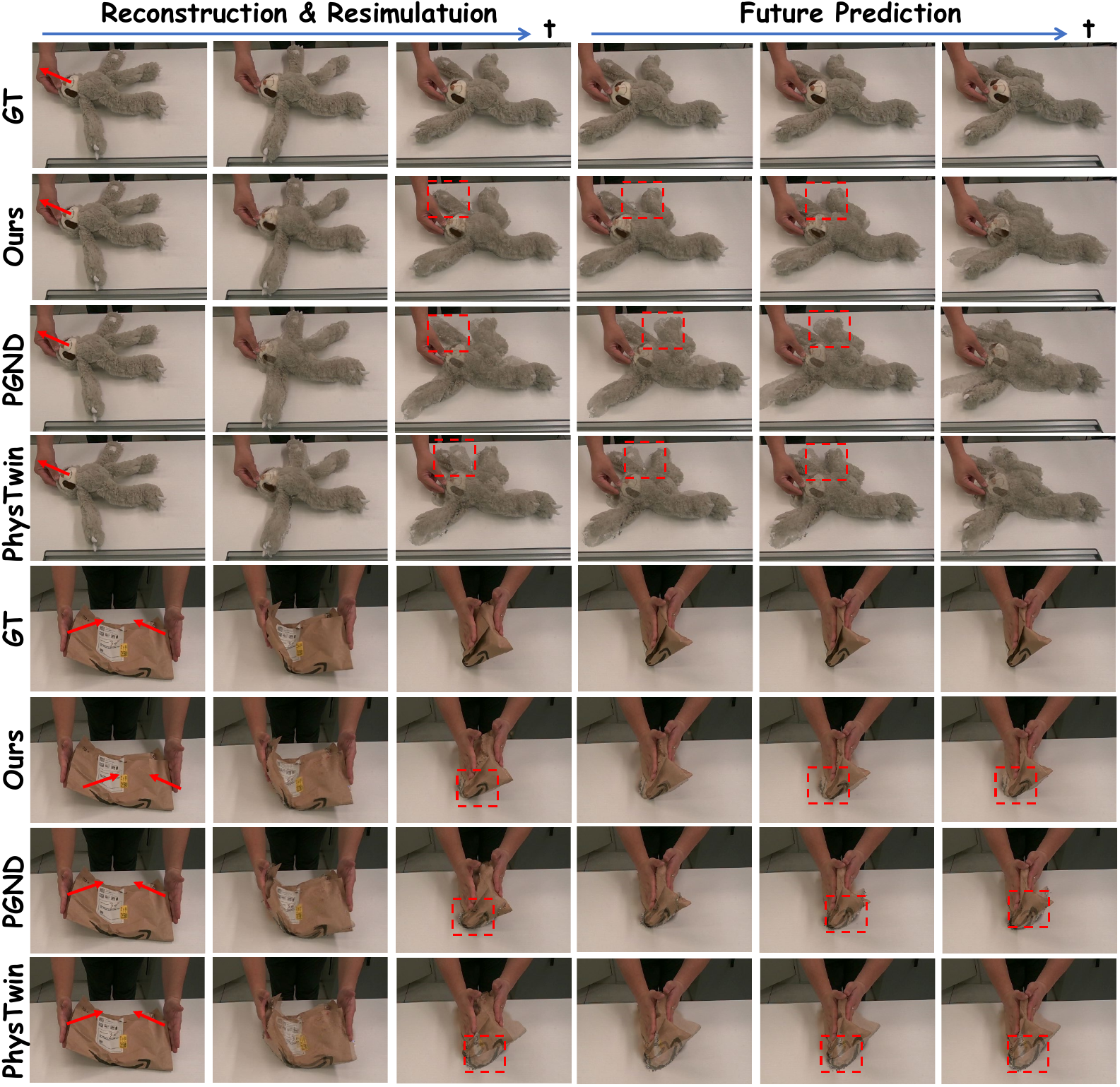}
\caption{Additional qualitative comparisons on PhysTwin for reconstruction and resimulation (left) and future prediction (right). Red boxes highlight local deformation differences.}
\label{fig:phystwin_supp_qualitative}
\end{figure}

This section supplements the within-domain results in Table~\ref{tab:phystwin_main} with additional qualitative comparisons on PhysTwin. Figure~\ref{fig:phystwin_supp_qualitative} shows a plush object undergoing localized manipulation and a deformable package compressed from both sides. The left panels evaluate reconstruction and resimulation under recorded controls, while the right panels examine future prediction beyond the observed interval. These examples illustrate how local manipulation produces different responses across the object.

\textbf{Reconstruction and Resimulation.}
\themodel more closely reproduces the observed configurations in both examples. For the plush object, the highlighted limb region remains better aligned with the ground truth, whereas the baselines exhibit more noticeable deviations in its position and shape. This distinction matters when nearby object regions undergo different motions: spatial proximity alone does not imply that they should deform together. The package example tests a complementary behavior, where two-sided compression produces localized folds and changes in the lower contour. Our prediction more closely follows this contour, while the baselines show larger distortions in the highlighted region. Together, these observations support improved modeling of spatially distributed deformation, beyond reproducing the overall direction of manipulation.

\textbf{Future Prediction.}
The differences persist as prediction extends beyond the observed interval. \themodel better preserves the configuration of the highlighted plush limb and the folded lower region of the package, while the baseline predictions depart more visibly from the corresponding ground-truth shapes. The agreement across both tasks suggests that the tracking improvements reflect more faithful deformation dynamics, rather than only a closer fit to observed states. In particular, the reduced local distortions are consistent with better discrimination between spatial proximity and deformation-relevant coupling, addressing the local-interaction ambiguity illustrated in Fig.~\ref{fig:main_fig}.

\subsection{Per-Category Results on Held-Out Interaction Prediction}
\label{app:pgnd_rollout}

\begin{table*}[t]
\centering
\caption{Quantitative comparison with PGND on unseen interaction prediction across six deformable object categories. Best results are shown in \textbf{bold}.}
\label{tab:pgnd_per_category}

\small
\setlength{\tabcolsep}{4pt}
\renewcommand{\arraystretch}{1.12}

\begin{tabularx}{\textwidth}{@{}ll*{4}{>{\raggedleft\arraybackslash}X}r*{3}{>{\raggedleft\arraybackslash}X}@{}}
\toprule
\textbf{Object} & \textbf{Method}
& MDE $\downarrow$
& CD $\downarrow$
& EMD $\downarrow$
& IoU $\uparrow$
& Boundary F $\uparrow$
& PSNR $\uparrow$
& SSIM $\uparrow$
& LPIPS $\downarrow$ \\
\midrule

& PGND
& 45.80 & 43.07 & 22.05 & 65.40
& \textbf{44.22} & 23.86 & 0.916 & 0.108 \\
\rowcolor{oursbg}
\cellcolor{white}\multirow{-2}{*}{Cloth}
& \themodel
& \textbf{44.15} & \textbf{38.67} & \textbf{18.45} & \textbf{66.13}
& 43.95 & \textbf{23.92} & \textbf{0.919} & \textbf{0.106} \\
\addlinespace[5pt]

& PGND
& 47.52 & 47.47 & 26.35 & 25.43
& 46.68 & 23.34 & 0.975 & 0.042 \\
\rowcolor{oursbg}
\cellcolor{white}\multirow{-2}{*}{Rope}
& \themodel
& \textbf{22.49} & \textbf{20.63} & \textbf{10.87} & \textbf{32.24}
& \textbf{57.50} & \textbf{23.39} & \textbf{0.977} & \textbf{0.034} \\
\addlinespace[5pt]

& PGND
& 45.07 & 34.36 & 18.03 & \textbf{58.90}
& 47.20 & 23.40 & 0.932 & 0.069 \\
\rowcolor{oursbg}
\cellcolor{white}\multirow{-2}{*}{Plush}
& \themodel
& \textbf{38.71} & \textbf{29.89} & \textbf{14.70} & 58.62
& \textbf{50.30} & \textbf{23.44} & \textbf{0.935} & \textbf{0.067} \\
\addlinespace[5pt]

& PGND
& 21.34 & 26.43 & 15.40 & 84.66
& 70.28 & 25.92 & 0.929 & 0.087 \\
\rowcolor{oursbg}
\cellcolor{white}\multirow{-2}{*}{Box}
& \themodel
& \textbf{19.41} & \textbf{23.41} & \textbf{13.33} & \textbf{87.05}
& \textbf{76.35} & \textbf{25.95} & \textbf{0.936} & \textbf{0.079} \\
\addlinespace[5pt]

& PGND
& 16.24 & 21.12 & 9.61 & 77.32
& 57.76 & 24.57 & 0.934 & 0.064 \\
\rowcolor{oursbg}
\cellcolor{white}\multirow{-2}{*}{Bag}
& \themodel
& \textbf{13.97} & \textbf{19.42} & \textbf{8.57} & \textbf{80.39}
& \textbf{65.39} & \textbf{24.61} & \textbf{0.939} & \textbf{0.058} \\
\addlinespace[5pt]

& PGND
& 21.59 & 18.66 & 10.84 & 53.48
& 65.62 & 23.75 & 0.967 & 0.043 \\
\rowcolor{oursbg}
\cellcolor{white}\multirow{-2}{*}{Bread}
& \themodel
& \textbf{13.78} & \textbf{14.35} & \textbf{7.81} & \textbf{62.64}
& \textbf{75.73} & \textbf{23.88} & \textbf{0.973} & \textbf{0.033} \\

\bottomrule
\end{tabularx}
\end{table*}

Table~\ref{tab:pgnd_per_category} supplements the category-averaged temporal curves in Fig.~\ref{fig:pgnd_avg_metric} with detailed results across cloth, rope, plush, box, bag, and bread. Under the same held-out interaction protocol, \themodel achieves lower MDE, CD, and EMD in every category, showing that the geometric improvements extend across diverse deformable objects.

The largest motion-prediction gains occur on rope and bread, where MDE decreases by 52.7\% and 36.2\%, respectively, relative to PGND. The substantial improvement on rope is consistent with the qualitative comparison in Fig.~\ref{fig:pgnd_rollout_inst_rope}, where our prediction better follows the object's evolving configuration. Improvements on cloth are more modest, while the remaining categories also show consistent reductions across all three particle-space errors. Thus, the aggregate advantage reflects broadly improved geometric prediction, although its magnitude varies across categories.

The visual metrics provide complementary evidence: \themodel improves SSIM and LPIPS in all six categories, and achieves higher IoU and Boundary F in five categories each. The exceptions are small decreases in plush IoU and cloth Boundary F, indicating that lower particle-space errors do not always translate into improvements in every silhouette metric. PSNR differences are comparatively small, including a slight decrease on rope. These results suggest that the clearest benefits lie in more accurate particle motion and improved structural and perceptual fidelity, rather than large changes in pixel-level reconstruction quality.

\textbf{Single-Scene Generalization.}
\label{app:phystwin_cross_action}
\begin{figure}[t] 
\centering
\includegraphics[width=1.0\linewidth]{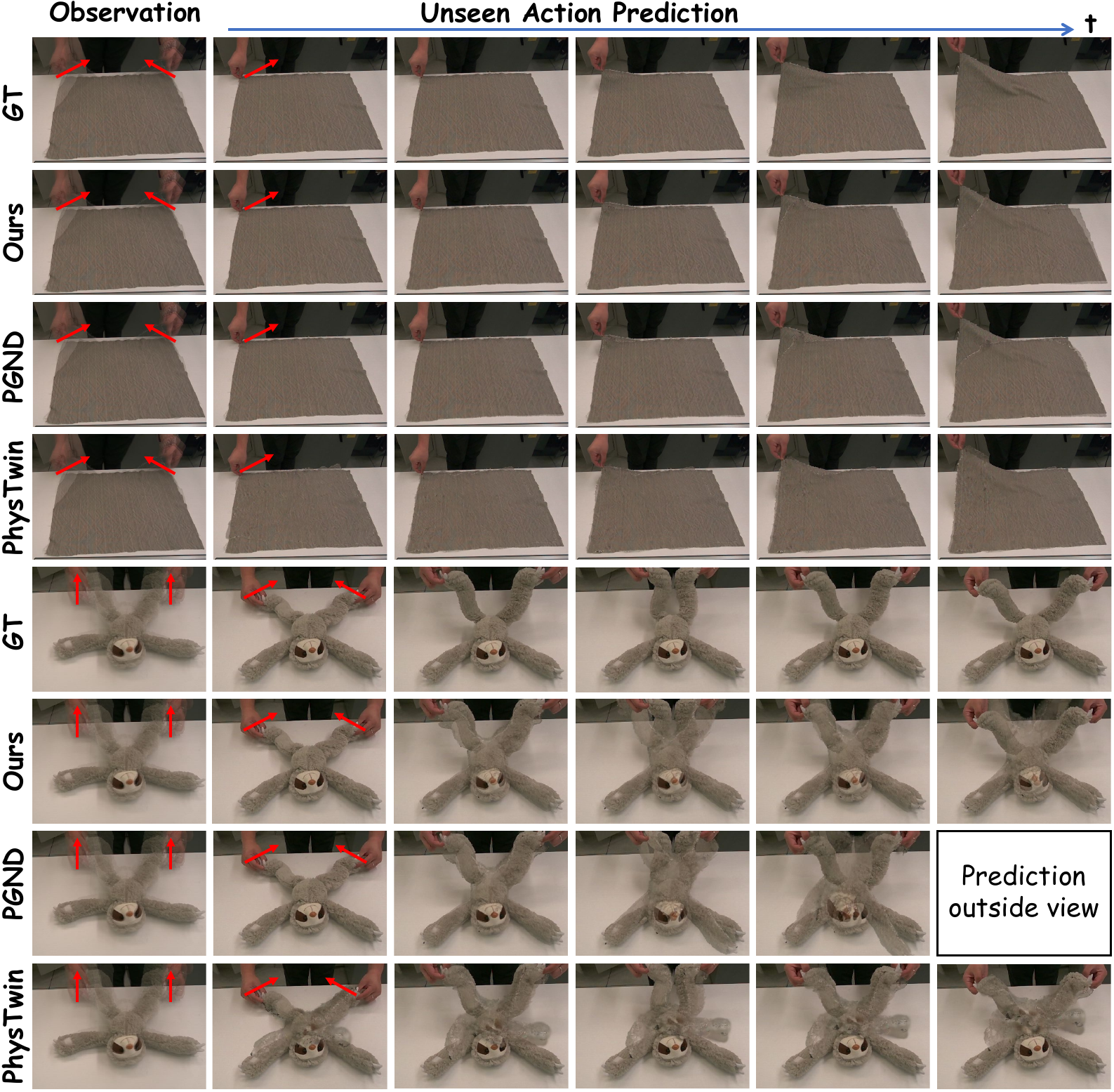}
\caption{Qualitative Results on Cross-Action Transfer on the Same Object.}
\label{fig:phystwin_action_genralize}
\end{figure}

\begin{table*}[t]
\centering
\caption{Per-scene results on cross-action transfer. Each model is transferred from the source interaction to the target interaction without target-specific training. Best results within each transfer pair are shown in \textbf{bold}.}
\label{tab:phystwin_cross_action_details}

\small
\setlength{\tabcolsep}{5pt}
\renewcommand{\arraystretch}{1.12}

\begin{tabularx}{\textwidth}{@{}l*{6}{>{\raggedleft\arraybackslash}X}@{}}
\toprule
\textbf{Method}
& CD $\downarrow$
& Track $\downarrow$
& IoU $\uparrow$
& PSNR $\uparrow$
& SSIM $\uparrow$
& LPIPS $\downarrow$ \\
\midrule

\multicolumn{7}{@{}l}{\textit{Cloth 3: Double lift $\rightarrow$ Single clift}} \\
\addlinespace[2pt]
PhysTwin
& \textbf{0.010} & 0.029 & 88.6
& 24.310 & 0.877 & 0.126 \\
PGND
& 0.011 & 0.023 & 94.9
& 27.932 & 0.882 & 0.100 \\
\rowcolor{oursbg}
\themodel
& 0.011 & \textbf{0.022} & \textbf{95.3}
& \textbf{28.543} & \textbf{0.886} & \textbf{0.047} \\
\midrule

\multicolumn{7}{@{}l}{\textit{Sloth: Double lift $\rightarrow$ Double stretch}} \\
\addlinespace[2pt]
PhysTwin
& 0.051 & \textbf{0.063} & 35.4
& 19.662 & \textbf{0.899} & 0.132 \\
PGND
& 0.648 & 0.775 & 23.6
& 19.938 & 0.889 & 0.156 \\
\rowcolor{oursbg}
\themodel
& \textbf{0.034} & 0.089 & \textbf{49.4}
& \textbf{20.640} & 0.887 & \textbf{0.121} \\
\midrule

\multicolumn{7}{@{}l}{\textit{Zebra: Double stretch $\rightarrow$ Double lift}} \\
\addlinespace[2pt]
PhysTwin
& 0.045 & 0.066 & 35.4
& 21.063 & 0.924 & 0.100 \\
PGND
& 0.018 & 0.024 & 56.5
& 23.419 & 0.935 & 0.062 \\
\rowcolor{oursbg}
\themodel
& \textbf{0.015} & \textbf{0.024} & \textbf{67.0}
& \textbf{24.108} & \textbf{0.939} & \textbf{0.050} \\
\bottomrule
\end{tabularx}
\end{table*}

This section expands the averaged cross-action results in Table~\ref{tab:phystwin_transfer_mean} into three individual transfer pairs: double lift to single lift on Cloth 3, double lift to double stretch on Sloth, and double stretch to double lift on Zebra. Table~\ref{tab:phystwin_cross_action_details} reports per-scene performance, while Fig.~\ref{fig:phystwin_action_genralize} visualizes the cloth and sloth transfers. Each model reuses its source dynamics parameters without training on the target interaction.

\textbf{Per-Scene Quantitative Results.}
The improvements vary with the change in manipulation. On cloth, \themodel achieves the lowest tracking error and the best image-space scores, while PhysTwin retains a slightly lower CD. The lifting-to-stretching transfer on sloth produces a larger performance gap: \themodel reduces CD by approximately 33\% relative to PhysTwin and improves IoU by 14.0 percentage points, although PhysTwin achieves better tracking and SSIM. On zebra, the advantage extends to the reverse action change, with \themodel obtaining the best results across the reported metrics. Its tracking error is close to PGND, but IoU improves by 10.5 percentage points, indicating a clearer benefit in preserving the overall object configuration. These results show that the average gains are accompanied by improvements across distinct action changes, with some metric-specific trade-offs.

\textbf{Qualitative Comparisons.}
Figure~\ref{fig:phystwin_action_genralize} illustrates how these differences appear during prediction. On cloth, \themodel more closely reproduces the asymmetric deformation caused by switching from double lifting to single lifting, particularly around the raised region and the object boundary at later timesteps. On sloth, stretching requires a different coordination of the torso and limbs from the source lifting motion. Our predictions better preserve their relative configuration, whereas PGND develops substantial distortion and eventually moves outside the displayed view. PhysTwin remains within view but exhibits visible differences in the central body shape. These examples complement the quantitative results by showing how inaccurate transfer changes the spatial deformation induced by the target controls.

\textbf{Category-Level Generalization.}
\label{app:phystwin_cross_object}
\begin{figure}[t] 
\centering
\includegraphics[width=1.0\linewidth]{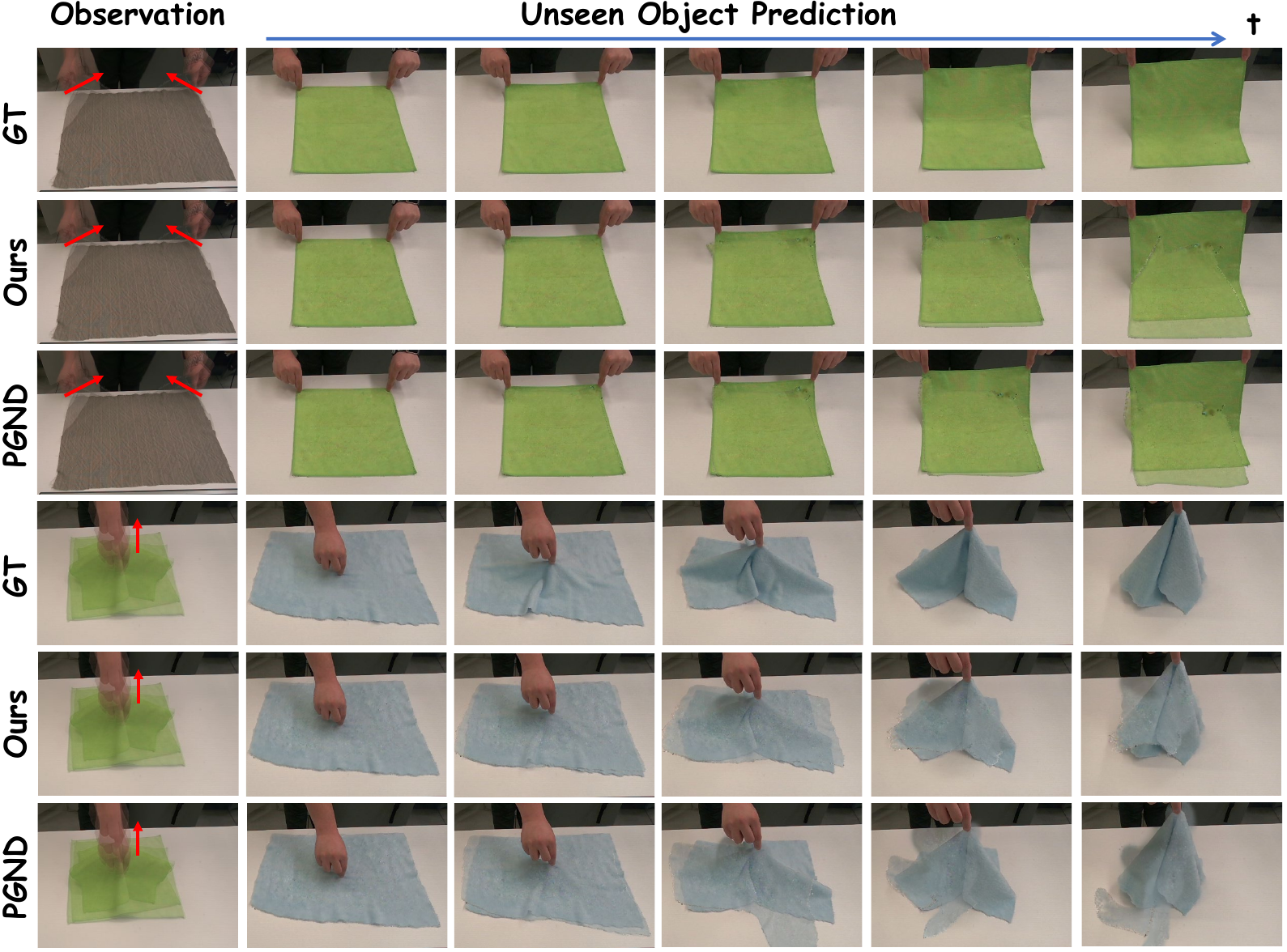}
\caption{Qualitative Results on Cross-Object Transfer under Matched Action Types.}
\label{fig:phystwin_object_genralize}
\end{figure}

\begin{table*}[t]
\centering
\caption{Per-scene results on cross-object transfer under matched action types. Best displayed results within each transfer pair are shown in \textbf{bold}.}
\label{tab:phystwin_cross_object_details}

\small
\setlength{\tabcolsep}{5pt}
\renewcommand{\arraystretch}{1.12}

\begin{tabularx}{\textwidth}{@{}l*{6}{>{\raggedleft\arraybackslash}X}@{}}
\toprule
\textbf{Method}
& CD $\downarrow$
& Track $\downarrow$
& IoU $\uparrow$
& PSNR $\uparrow$
& SSIM $\uparrow$
& LPIPS $\downarrow$ \\
\midrule

\multicolumn{7}{@{}l}{\textit{Single lift: Cloth 1 $\rightarrow$ Cloth 4}} \\
\addlinespace[2pt]
PGND
& 0.040 & 0.053 & 62.4
& 20.332 & 0.922 & 0.107 \\
\rowcolor{oursbg}
\themodel
& \textbf{0.023} & \textbf{0.029} & \textbf{76.9}
& \textbf{23.066} & \textbf{0.939} & \textbf{0.081} \\
\midrule

\multicolumn{7}{@{}l}{\textit{Double lift: Cloth 3 $\rightarrow$ Cloth 1}} \\
\addlinespace[2pt]
PGND
& \textbf{0.012} & 0.021 & 81.9
& 25.715 & \textbf{0.945} & 0.065 \\
\rowcolor{oursbg}
\themodel
& \textbf{0.012} & \textbf{0.016} & \textbf{82.6}
& \textbf{25.945} & 0.944 & \textbf{0.064} \\
\bottomrule
\end{tabularx}
\end{table*}

This section supplements the averaged cross-object results in Table~\ref{tab:phystwin_transfer_mean} with two individual transfers: single lifting from Cloth 1 to Cloth 4 and double lifting from Cloth 3 to Cloth 1. Table~\ref{tab:phystwin_cross_object_details} reports per-scene performance, and Fig.~\ref{fig:phystwin_object_genralize} provides corresponding qualitative comparisons. The learned dynamics are reused on the target object without target-specific training. PhysTwin is excluded because transferring its object-specific physical representation requires additional correspondence or adaptation beyond this evaluation protocol.

\textbf{Per-Scene Quantitative Results.}
The gains are most pronounced in the single-lift transfer, where \themodel improves all reported metrics over PGND. Tracking error decreases by approximately 45\%, while IoU increases by 14.5 percentage points, indicating better agreement in both particle motion and projected object shape. The double-lift transfer shows a smaller but consistent advantage in tracking, with an approximately 24\% reduction in error despite nearly identical CD. Rendering improvements are modest in this case, and PGND retains a slight advantage in SSIM. These results show that the average improvement is larger for the single-lift pair, while more accurate motion tracking is observed in both transfers.

\textbf{Qualitative Comparisons.}
Figure~\ref{fig:phystwin_object_genralize} illustrates these differences in the evolving cloth configurations. In the double-lift example, both methods reproduce the overall upward motion, but \themodel more closely follows the target cloth's lower boundary. The distinction is clearer in the single-lift example, where localized pulling produces folds and a progressively narrowing silhouette. Our prediction better captures this deformation, whereas PGND retains a broader lower region and deviates more visibly from the observed shape at later timesteps. These examples suggest that the tracking gains are accompanied by better preservation of the target object's distributed deformation under the transferred dynamics.

\subsection{Zero-Shot Cross-Dataset Transfer}
\label{app:cross_dataset_transfer}

This section supplements the averaged cross-dataset results in Table~\ref{tab:cross_dataset_transfer} with separate evaluations on rope, cloth, and sloth. Models trained on PGND robot interactions are transferred to human-manipulated PhysTwin sequences after coordinate and temporal alignment. All learned dynamics parameters remain frozen, with no target-domain dynamics fitting. Table~\ref{tab:cross_dataset_transfer_details} reports the individual results, assessing whether motion prediction remains accurate across changes in the data source and manipulation setting.

\begin{table}[t]
    \centering
    \caption{Per-scene results on zero-shot transfer from PGND to PhysTwin. Best results within each scene are shown in \textbf{bold}.}
    \label{tab:cross_dataset_transfer_details}
    \small
    \setlength{\tabcolsep}{4pt}
    \renewcommand{\arraystretch}{1.08}
    \begin{tabularx}{\linewidth}{@{}>{\raggedright\arraybackslash}X*{6}{>{\centering\arraybackslash}X}@{}}
        \toprule
        \textbf{Method} & CD $\downarrow$ & Track $\downarrow$ & IoU $\uparrow$ & PSNR $\uparrow$ & SSIM $\uparrow$ & LPIPS $\downarrow$ \\
        \midrule
        \multicolumn{7}{@{}l}{\textit{Rope: robot $\rightarrow$ human manipulation}} \\
        PGND & 0.014 & 0.029 & 39.70 & 26.132 & 0.976 & 0.035 \\
        \rowcolor{oursbg}
        \textbf{\themodel} & \textbf{0.013} & \textbf{0.017} & \textbf{54.00} & \textbf{27.738} & \textbf{0.977} & \textbf{0.030} \\
        \midrule
        \multicolumn{7}{@{}l}{\textit{Cloth: robot $\rightarrow$ human manipulation}} \\
        PGND & 0.021 & 0.032 & 76.60 & 25.130 & 0.926 & 0.088 \\
        \rowcolor{oursbg}
        \textbf{\themodel} & \textbf{0.017} & \textbf{0.022} & \textbf{78.50} & \textbf{25.561} & \textbf{0.927} & \textbf{0.085} \\
        \midrule
        \multicolumn{7}{@{}l}{\textit{Sloth: robot $\rightarrow$ human manipulation}} \\
        PGND & 0.021 & 0.039 & \textbf{64.54} & \textbf{23.242} & \textbf{0.923} & \textbf{0.087} \\
        \rowcolor{oursbg}
        \textbf{\themodel} & \textbf{0.019} & \textbf{0.030} & 63.13 & 23.014 & 0.919 & 0.092 \\
        \bottomrule
    \end{tabularx}
\end{table}

\textbf{Per-Scene Quantitative Results.}
The strongest improvement occurs on rope, where \themodel reduces tracking error by approximately 41\% relative to PGND. The accompanying improvement in object overlap indicates that more accurate point motion also yields closer agreement with the projected rope configuration. On cloth, tracking error decreases by approximately 31\%, with consistent but smaller gains in rendered appearance. These results suggest that the motion-prediction advantage persists across both elongated and sheet-like objects under human manipulation.

The sloth transfer presents a more mixed outcome. Although \themodel reduces tracking error by approximately 23\% and achieves lower CD, PGND retains a slight advantage across the image-space metrics. Thus, improved particle motion does not translate uniformly into better rendered appearance in this setting. Overall, tracking improves in all three scenes, while the visual benefits are concentrated on rope and cloth, revealing both the utility and the remaining limitations of transferring the learned dynamics without target-specific adaptation.

\subsection{Additional Ablation Analysis}
\label{app:ablation_details}

We provide further details on the ablations in Fig.~\ref{fig:ablation}, conducted on \texttt{double\_stretch\_zebra}. All variants use the same evaluation protocol and are compared at their individually selected checkpoints. The study examines adaptive local relations, global communication, and geometry-conditioned redistribution, including a joint-removal baseline and a distance-based weight-assignment variant.

\textbf{Variant Definitions.}
The full model combines adaptive interaction relations with a geometry-conditioned global token. The \textit{radius-local} variant replaces adaptive relations with a spatial radius graph and removes the global pathway, while \textit{adaptive-local} retains adaptive relations but removes the global token. The \textit{radius-global} variant uses radius-based relations together with the geometry-conditioned global pathway. Both radius variants also remove the learned relation-feature injection. The \textit{plain-global} variant retains token initialization, aggregation, and updating, but omits the six explicit relative-geometry inputs during redistribution to object particles. Finally, \textit{distance-rank} reassigns learned object--object weights according to pair distance, assigning larger weights to nearer pairs. It preserves the candidate construction, learned relation embeddings, weight values, and original relation-retention state; control edges remain unchanged.

\textbf{Adaptive Interaction Relations.}
Replacing adaptive relations with a radius graph reduces IoU from 84.6\% to 81.4\%, despite retaining the global pathway. The degradation suggests that shared object-level information benefits from deformation-dependent local communication. The distance-ranking comparison further examines how interaction strengths are assigned: the full model reduces tracking error by approximately 16\% relative to this variant. Since distance ranking retains learned weight values and relation features, this result supports the usefulness of the learned association between particle pairs and their interaction strengths, beyond proximity-based ordering.

\textbf{Global Context Integration.}
Removing the global token increases tracking error from 6.6 to 15.9\,mm. Equivalently, incorporating the global pathway reduces this error by approximately 58\%, with corresponding improvements in object overlap and rendering quality. This is the largest degradation among the variants that remove adaptive relations or global communication individually, highlighting the importance of object-level information for particle updates. Jointly removing both mechanisms produces the largest errors overall, indicating that the local radius graph alone is insufficient to match the full model in this setting.

\textbf{Geometry-Conditioned Redistribution.}
The plain-global variant retains particle features and the shared token, allowing particle-dependent responses even without the explicit geometric descriptor. Nevertheless, omitting this descriptor reduces IoU by 5.1 percentage points, while CD and tracking error change more modestly. The stronger effect on object overlap suggests that relative object--control geometry helps spatially localize the global context, improving the projected configuration beyond what is obtained from the remaining latent features alone.

\textbf{Scope of the Comparison.}
These results characterize prediction performance on one scene and training seed. The variants differ in parameter count, so the comparison does not isolate architectural effects under matched parameter counts. In particular, the radius variants replace both relation inference and relation-feature injection, while distance ranking changes weight assignment rather than removing all relation learning. Long-horizon stability and transferability are evaluated separately in the main experiments.

\section{Limitations and Broader Impact}
\label{app:limitations_impact}

\subsection{Limitations}
Although \themodel improves dynamics prediction and transfer across interaction conditions, it relies on reconstructed particle states and tracked control trajectories. Errors in reconstruction, occlusion handling, or control tracking can therefore affect the predicted deformation. Moreover, the adaptive interaction graph and global context improve empirical rollout accuracy but do not explicitly enforce physical conservation laws or guarantee stability over arbitrarily long horizons. Our generalization results are also limited to the evaluated objects and manipulation settings; they do not establish robustness to substantially different material properties, contact regimes, or topology-changing events such as tearing. Extending the evaluation to these conditions and to closed-loop manipulation remains future work.

\subsection{Broader Impact}
\themodel may support robotic manipulation of deformable objects by providing reusable dynamics predictions for simulation, planning, and policy evaluation. Its ability to transfer learned responses across objects and actions could reduce the need for repeated physical data collection and object-specific model fitting. However, inaccurate predictions under unfamiliar conditions could lead to failed manipulation or damage to objects and equipment if used directly for control. Deployment should therefore include validation under the intended operating conditions, monitoring of prediction errors, and appropriate control safeguards. The present results demonstrate predictive performance on recorded interactions rather than validated safety or reliability in autonomous robotic operation.

\end{document}